\documentclass{article}

\usepackage{microtype}
\usepackage{graphicx}
\usepackage{subcaption}
\usepackage{booktabs} 
\usepackage{multirow}
\usepackage{algorithmic}

\usepackage{hyperref}

\newcommand{\x}{\textbf{x}}
\newcommand{\vel}{\textbf{v}}

\newcommand{\Zero}{\textbf{O}}
\newcommand{\E}{\mathbb{E}}
\newcommand{\mean}{\boldsymbol{\mu}}
\newcommand{\cond}{{\text{cond}}}
\newcommand{\uncond}{{\text{uncond}}}
\newcommand{\One}{\textbf{I}}
\newcommand{\norm}[2][]{\left\Vert #2 \right\Vert_{#1}}
\newcommand{\mtd}{\textbf{MAMBO-G}}
\newcommand{\mtdbk}{\textbf{MAMBO-G}~}
\newcommand{\Data}{\mathcal X}

\usepackage[preprint]{icml2026}

\usepackage{amsmath}
\usepackage{amssymb}
\usepackage{mathtools}
\usepackage{amsthm}

\usepackage[capitalize,noabbrev]{cleveref}
\usepackage{enumitem}

\theoremstyle{plain}

\theoremstyle{definition}

\theoremstyle{remark}

\newcommand{\methodabbr}{\textbf{MAMBO-G}}
\usepackage{xspace}

\usepackage[textsize=tiny]{todonotes}

\usepackage{caption}

\icmltitlerunning{}

\newcommand{\teasercode}{
    \vspace{10pt}
    \begin{center}
    \captionsetup{type=figure}
      
      \begin{minipage}{0.49\textwidth}
          \centering
          
          \begin{minipage}{0.32\linewidth} \centering \scriptsize \textbf{CFG ($\mathbf{20}$ NFE)} \end{minipage}%
          \hfill
          \begin{minipage}{0.32\linewidth} \centering \scriptsize \textbf{CFG ($\mathbf{60}$ NFE)} \end{minipage}%
          \hfill
          \begin{minipage}{0.32\linewidth} \centering \scriptsize \textbf{\mtdbk ($\mathbf{20}$ NFE)} \end{minipage}
          \vspace{2pt}

          \includegraphics[width=0.32\linewidth]{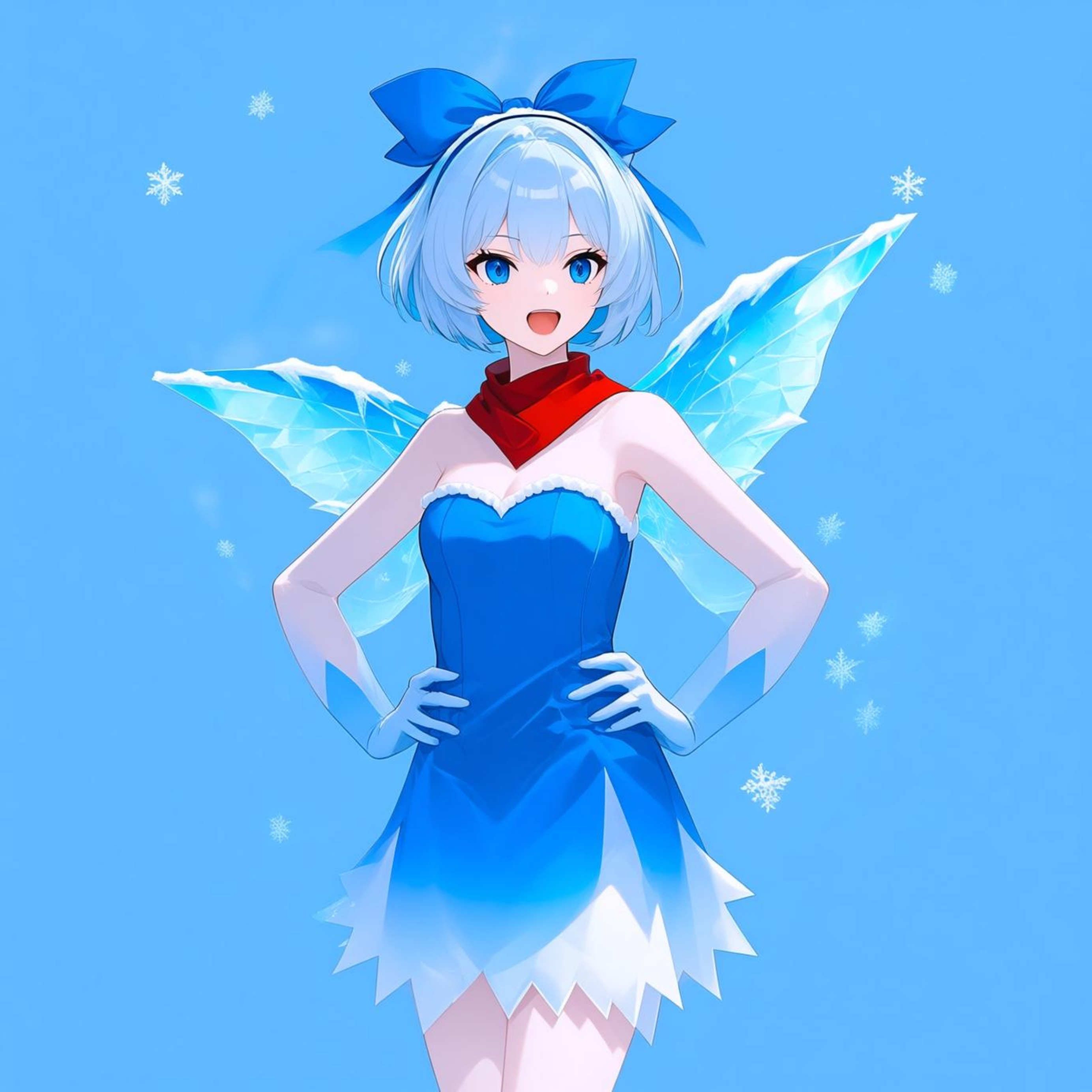}%
          \hfill
          \includegraphics[width=0.32\linewidth]{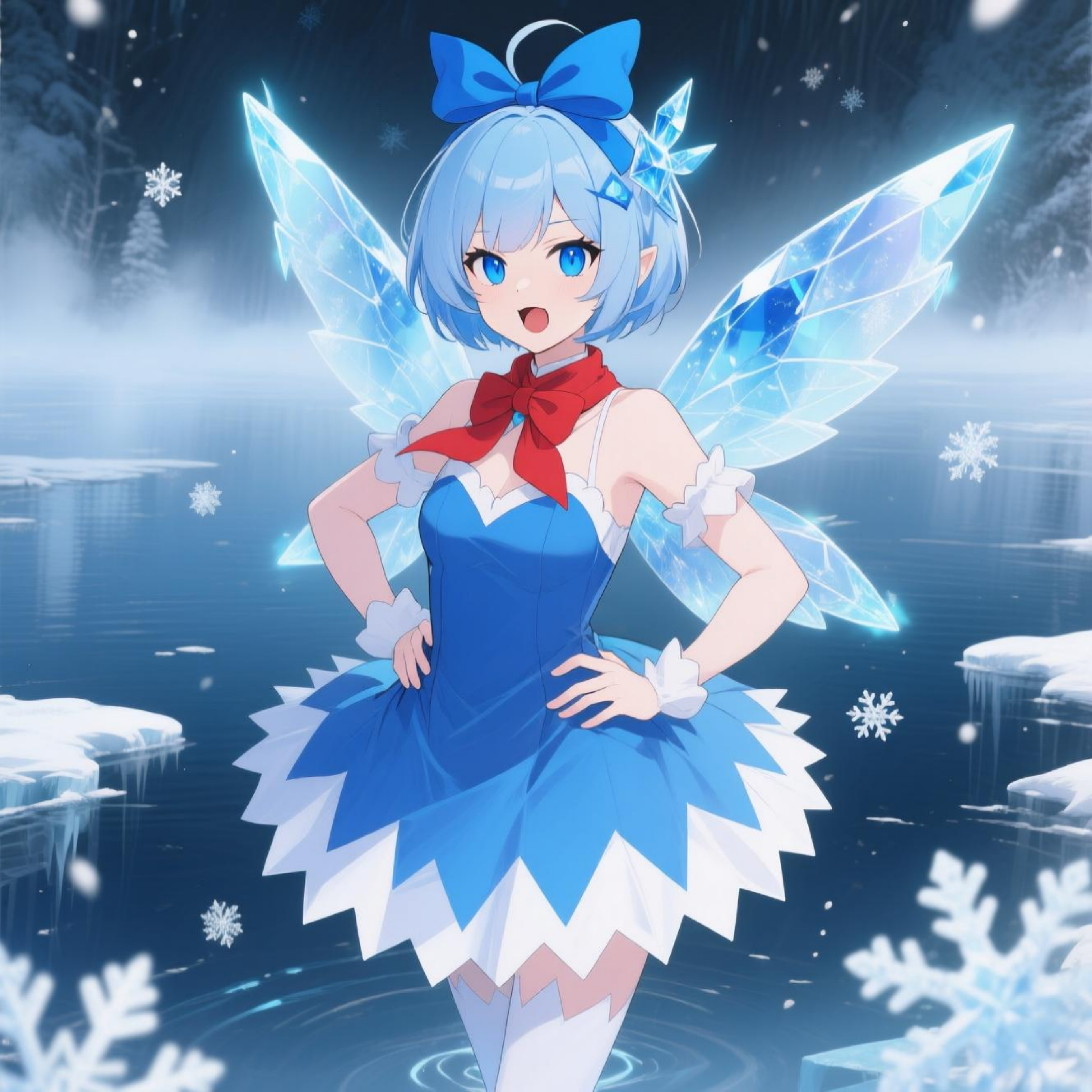}%
          \hfill
          \includegraphics[width=0.32\linewidth]{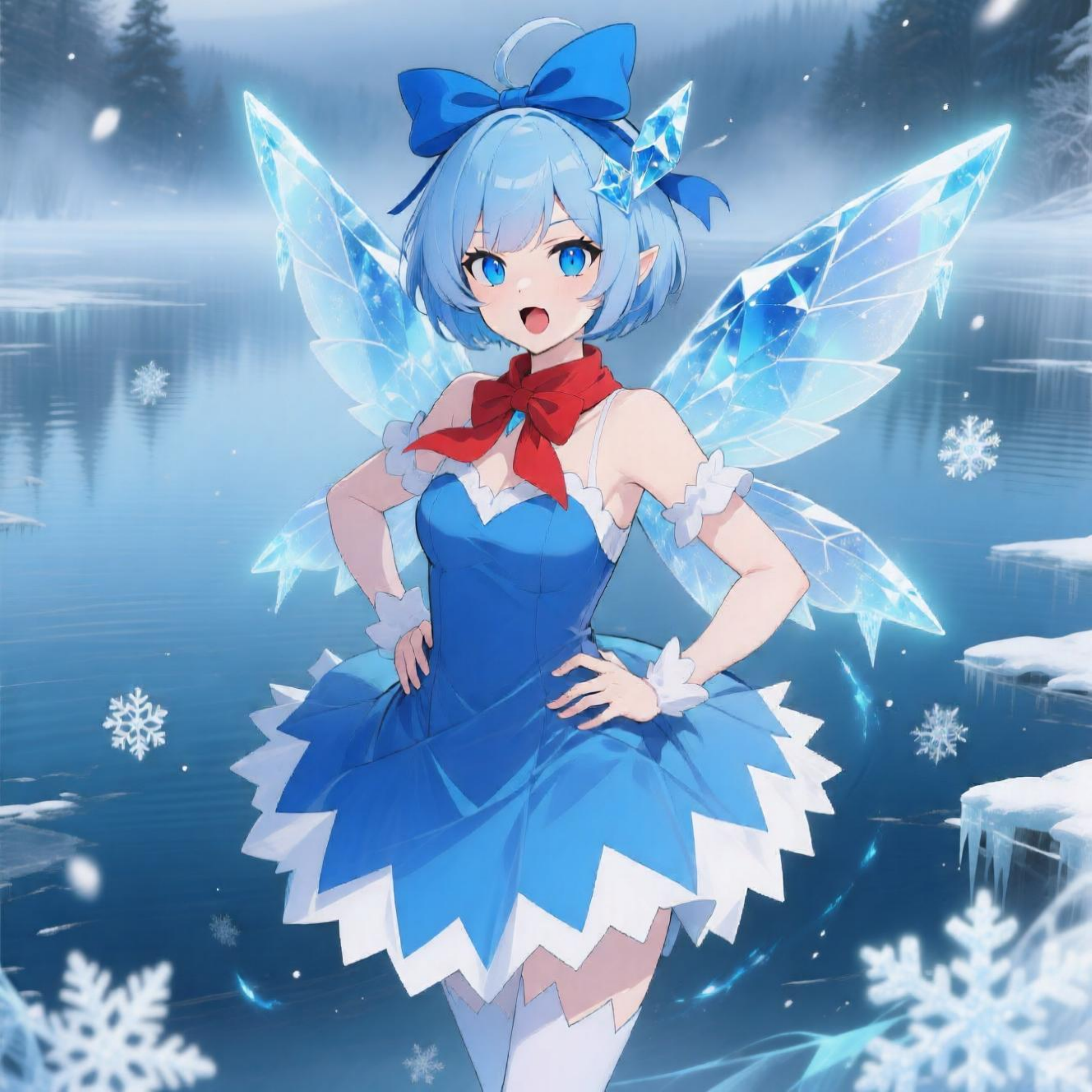}
          
          \vspace{2pt}
          \small (a)  Cirno from the Touhou Project (Seed 0 in Qwen-Image).
      \end{minipage}%
      \hfill
      \begin{minipage}{0.49\textwidth}
          \centering
         
          \begin{minipage}{0.32\linewidth} \centering \scriptsize \textbf{CFG ($\mathbf{20}$ NFE)} \end{minipage}%
          \hfill
          \begin{minipage}{0.32\linewidth} \centering \scriptsize \textbf{CFG ($\mathbf{60}$ NFE)} \end{minipage}%
          \hfill
          \begin{minipage}{0.32\linewidth} \centering \scriptsize \textbf{\mtdbk ($\mathbf{20}$ NFE)} \end{minipage}
          \vspace{2pt}

          \includegraphics[width=0.32\linewidth]{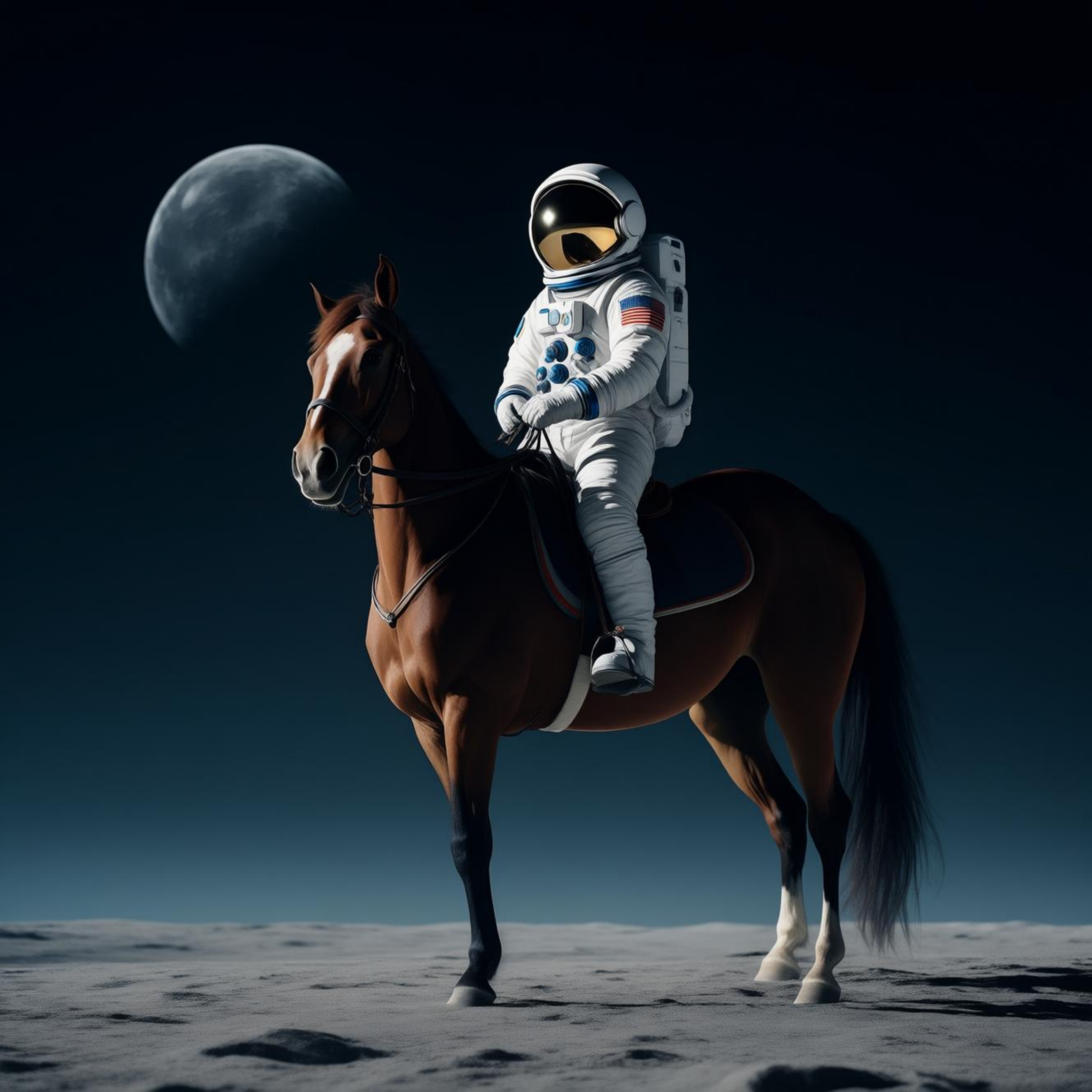}%
          \hfill
          \includegraphics[width=0.32\linewidth]{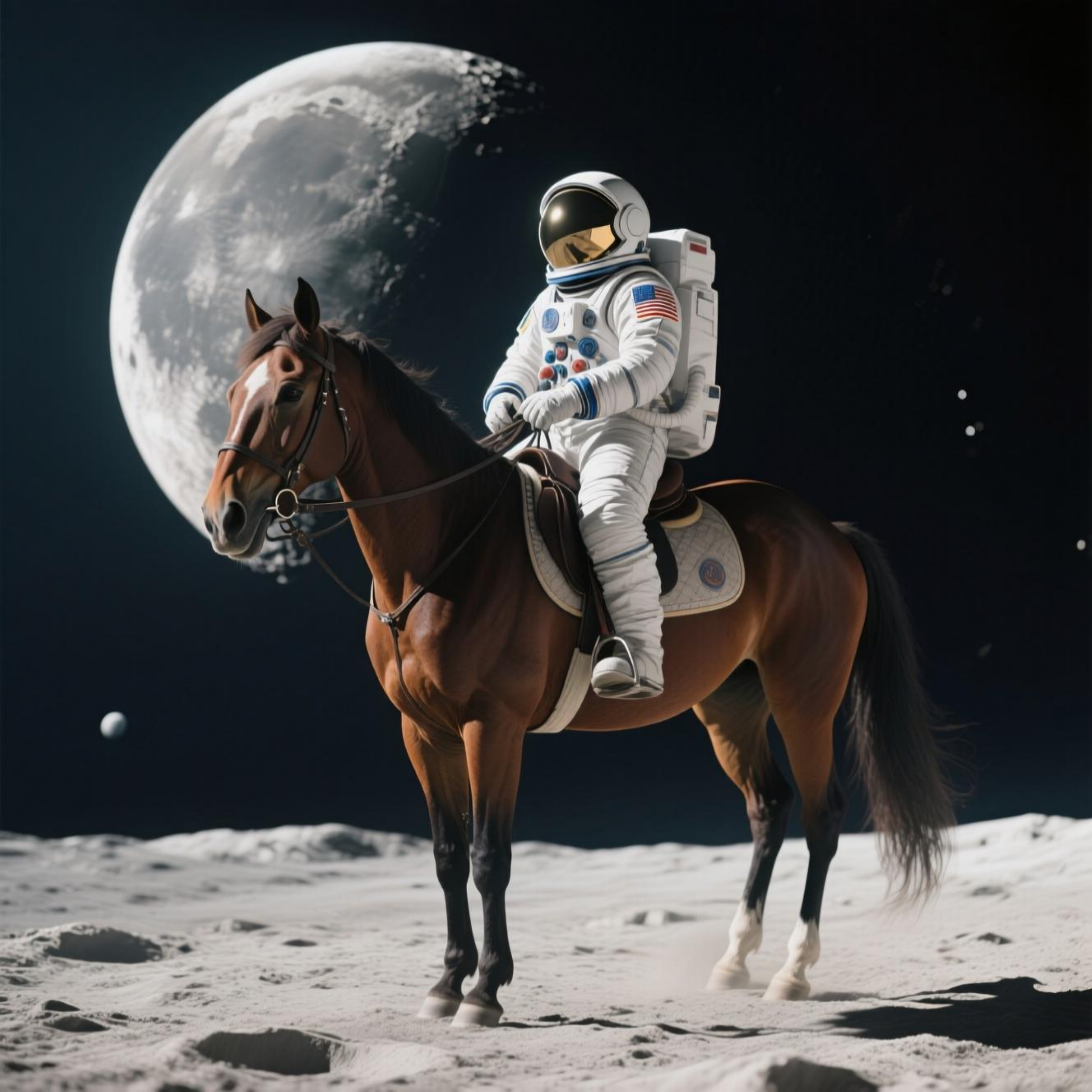}%
          \hfill
          \includegraphics[width=0.32\linewidth]{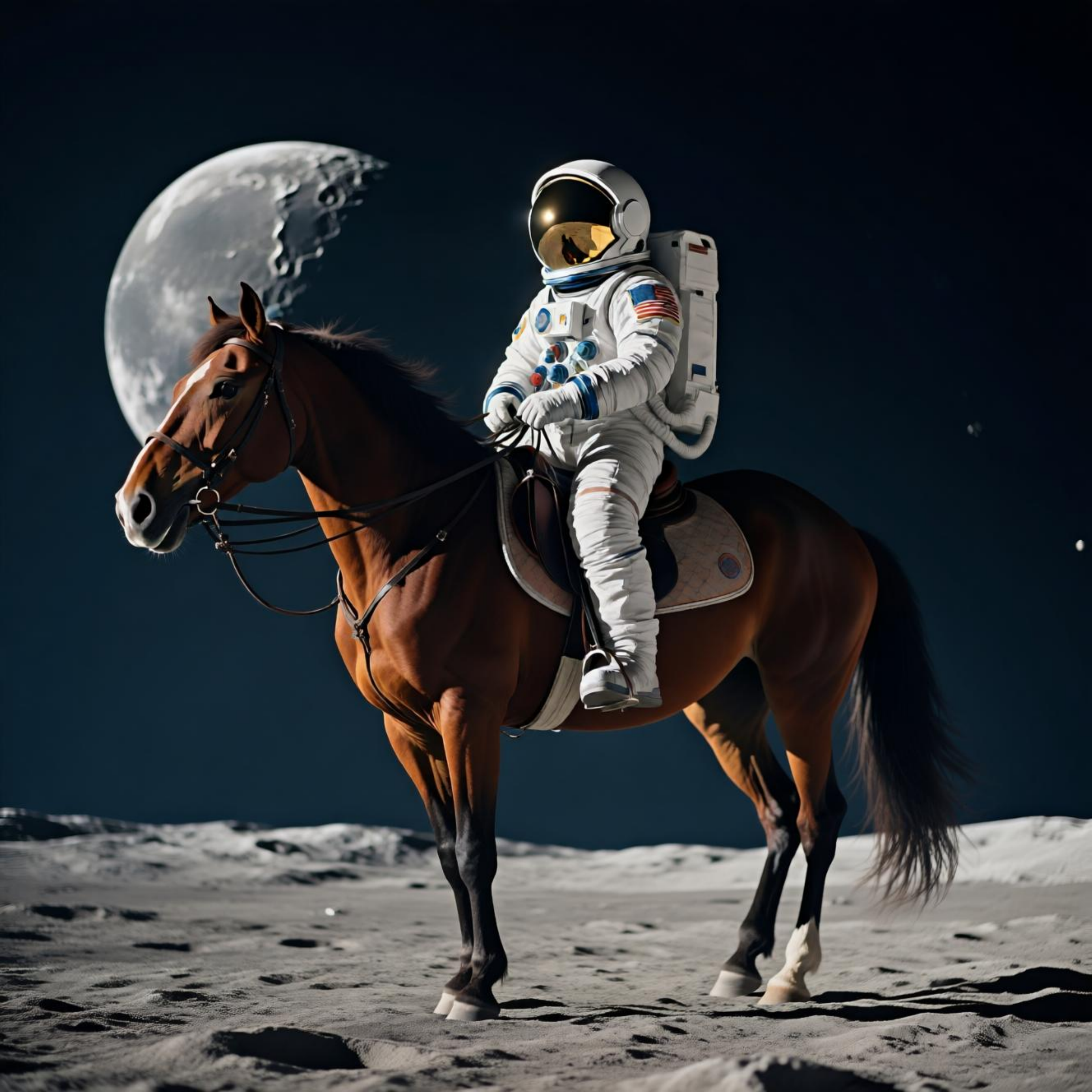}
          
          \vspace{2pt}
          \small (b) Classic astronaut riding a horse (Seed 0 in Qwen-Image).
      \end{minipage}
      \vspace{0.3cm}

      \begin{minipage}{1.0\textwidth}
          \centering
          \includegraphics[width=\linewidth]{fig_tab/film_strip_result_org.pdf}
          \\ \small (c) CFG(30NFE). Two cats fight on a spotlighted stage (Seed 0 in Wan2.2-5B, prompt from wan2.2 example).
      \end{minipage}
      
      \vspace{0.1cm}

      \begin{minipage}{1.0\textwidth}
          \centering
          \includegraphics[width=\linewidth]{fig_tab/film_strip_result_raag.pdf}
          \\ \small (d) \mtdbk(30NFE). Two cats fight on a spotlighted stage (Seed 0 in Wan2.2-5B).
      \end{minipage}
      \vspace{0.2cm}
      \captionof{figure}{\textbf{Superior efficiency of \mtdbk:} Our method achieves comparable quality to $60$-NFE ($30$-step) CFG image generation with only $20$ NFE ($10$ steps), demonstrating a $3.0\times$ speedup over the standard CFG sampling (guidance scale = 4.0). The examples we demonstrate are \textbf{not cherry-picked}, and the seeds are also marked. Specific prompts can be found in \Cref{app:prompts}.}
      \label{fig:display}
    \end{center}
    \vspace{20pt}
}

\begin{document}

\twocolumn[
  \icmltitle{MAMBO-G: Magnitude-Aware Mitigation for Boosted Guidance}



  \icmlsetsymbol{equal}{*}
  \icmlsetsymbol{corresponding}{$\dagger$}

  \icmlaffiliation{sjtu}{Shanghai Jiao Tong University}
  \icmlaffiliation{hku}{The University of Hong Kong}
  \icmlaffiliation{ustc}{University of Science and Technology of China}
  \icmlaffiliation{waterloo}{University of Waterloo}
  \icmlaffiliation{cas}{Chinese Academy of Sciences}
  \icmlaffiliation{zgc}{Zhongguancun Academy}

  \begin{icmlauthorlist}
    \icmlauthor{Shangwen Zhu}{equal,sjtu}
    \icmlauthor{Qianyu Peng}{equal,hku}
    \icmlauthor{Zhilei Shu}{equal,ustc}
    \icmlauthor{Yuting Hu}{sjtu}
    \icmlauthor{Zhantao Yang}{sjtu}
    \icmlauthor{Han Zhang}{sjtu}
    \icmlauthor{Zhao Pu}{sjtu}
    \icmlauthor{Andy Zheng}{waterloo}
    \icmlauthor{Xinyu Cui}{cas}
    \icmlauthor{Jian Zhao}{zgc}
    \icmlauthor{Ruili Feng}{corresponding,waterloo}
    \icmlauthor{Fan Cheng}{corresponding,sjtu}
  \end{icmlauthorlist}
  \icmlprintaffiliations 


  \icmlkeywords{}

  \vskip 0.3in

  \teasercode
]



\printAffiliationsAndNotice{%
  \begin{tabular}[t]{@{}r@{\,}l@{}}
    $^{*}$ & Equal contribution. \\
    $^{\dagger}$ & Corresponding authors.
  \end{tabular}%
}  

\begin{abstract}
  High-fidelity text-to-image and text-to-video generation typically relies on Classifier-Free Guidance (CFG), but achieving optimal results often demands computationally expensive sampling schedules. In this work, we propose \textbf{MAMBO-G}, a training-free acceleration framework that significantly reduces computational cost by dynamically optimizing guidance magnitudes. We observe that standard CFG schedules are inefficient, applying disproportionately large updates in early steps that hinder convergence speed. \textbf{MAMBO-G} mitigates this by modulating the guidance scale based on the update-to-prediction magnitude ratio, effectively stabilizing the trajectory and enabling rapid convergence. This efficiency is particularly vital for resource-intensive tasks like video generation. Our method serves as a universal plug-and-play accelerator, achieving up to 3$\times$ speedup on Stable Diffusion v3.5 (SD3.5) and 4$\times$ on Lumina. Most notably, \textbf{MAMBO-G} accelerates the 14B-parameter Wan2.1 video model by 2$\times$ while preserving visual fidelity, offering a practical solution for efficient large-scale video synthesis. 
  Our implementation follows mainstream open-source standards and is officially merged into the \href{https://github.com/huggingface/diffusers/pull/12862}{Diffusers} library, ensuring seamless plug-and-play integration with existing pipelines.
\end{abstract}

\section{Introduction}

Generative models have made significant progress in creating images and videos from text~\citep{song2020denoising,dhariwal2021diffusion,peebles2023scalable,ho2022video,esser2023structure}. A key technique they use is classifier-free guidance (CFG)~\citep{ho2022classifier}, which adjusts the model's output to better match the text prompt. However, using strong guidance can sometimes reduce stability, potentially leading to issues like oversaturated colors or unnatural structures~\citep{karczewski2024diffusion,sadat2024eliminating}. These problems are often more noticeable in modern, high-dimensional models, where using guidance scale without careful adjustment may strongly affect visual quality.

Recent studies have analyzed the stability and mechanisms of guidance strategies~\cite{lin2024common, wang2025foresight}. As models scale to latent spaces with millions of dimensions, they encounter challenges associated with high-dimensional spaces. In such environments, the initial noise magnitude naturally scales with dimensionality~\cite{debortoli2022riemannian}. Our analysis indicates that specifically at the initial timestep of generation, guidance update (difference between the conditional and unconditional model outputs) shares a similar direction across samples. In high-dimensional settings, forcing such a generic direction with a large guidance scale across diverse initial noises can destabilize the early generation trajectory, leading to severe overshooting and deviation from the realistic data distribution.

To address this, we propose \mtd. This method automatically adjusts the guidance scale by comparing it to the model's inherent denoising activity. When guidance is very strong relative to the denoising process, \mtdbk temporarily reduces the guidance scale, which helps keep the generation process stable in the early stages, while allowing a larger guidance scale later when the tones and structures of the image are clearer. The method is designed to be simple with almost no additional computational overhead, and to be compatible with various existing models and other CFG optimization strategies.

Our adaptive guidance schedule aims to accelerate conditional generation while maintaining sample quality, as suggested by metrics like ImageReward, CLIPScore, and vBench. Experiments indicate that \mtdbk can achieve faster inference on models such as SD3.5~\citep{esser2024scalingrectifiedflowtransformers}, Lumina~\citep{gao2024lumina}, and Wan2.1-14B~\citep{wan2.1} compared to baselines. Specifically, results show up to \textbf{3$\times$} acceleration on SD3.5, \textbf{4$\times$} on Lumina, and \textbf{2$\times$} on Wan2.1-14B, while achieving comparable or better performance than their slower baselines. The method also appears robust to hyperparameter settings and applicable across different model scales and domains. Moreover, \mtdbk is orthogonal to other guidance optimization methods, such as Guidance Rescale~\cite{lin2024common} and Adaptive Projection Guidance~\cite{sadat2024eliminating}, enabling seamless integration for cumulative benefits.

In summary, our main contributions are as follows:
\begin{itemize}
   \item We analyze the impact of early-step guidance in flow-based models, supported by theoretical motivation and empirical observations.
   \item We introduce a practical, magnitude-aware adaptive guidance schedule that aims to balance guidance scale across sampling steps.
   \item Through experiments on image and video generation, we demonstrate that our approach can facilitate speedups while maintaining the quality compared to standard CFG, with broad compatibility.
\end{itemize}

\label{sec:exp_video}
\begin{figure*}[t]
    \centering
    \begin{minipage}{\linewidth}
        \centering
        \begin{minipage}{0.24\linewidth} \centering \small Resolution: 256 \end{minipage}%
        \hfill
        \begin{minipage}{0.24\linewidth} \centering \small Resolution: 512 \end{minipage}%
        \hfill
        \begin{minipage}{0.24\linewidth} \centering \small Resolution: 768 \end{minipage}%
        \hfill
        \begin{minipage}{0.24\linewidth} \centering \small Resolution: 1024 \end{minipage}
    \end{minipage}
    \vspace{2pt}



    \begin{subfigure}{\linewidth}
        \centering
        \includegraphics[width=0.24\linewidth]{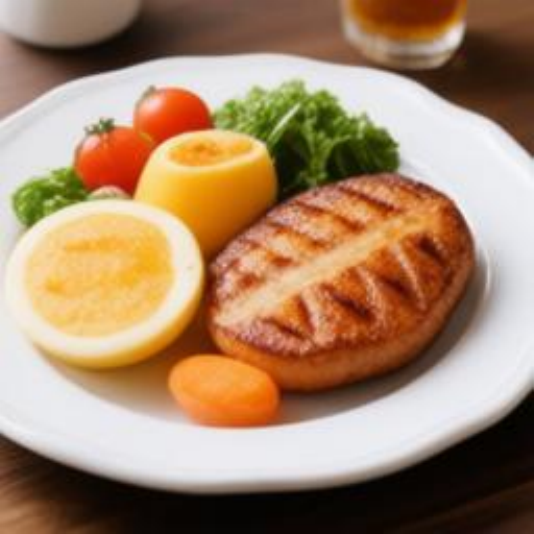}
        \hfill
        \includegraphics[width=0.24\linewidth]{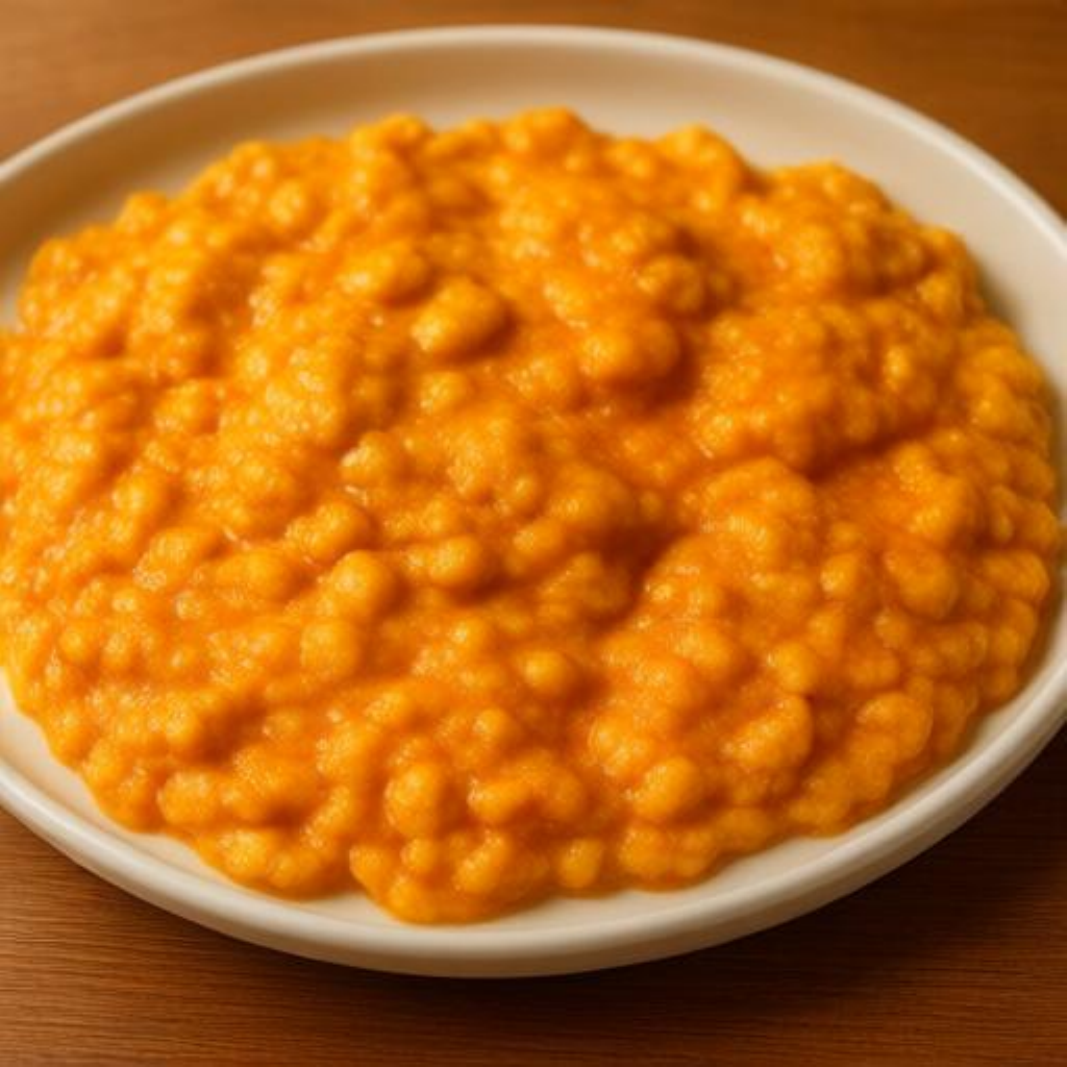}
        \hfill
        \includegraphics[width=0.24\linewidth]{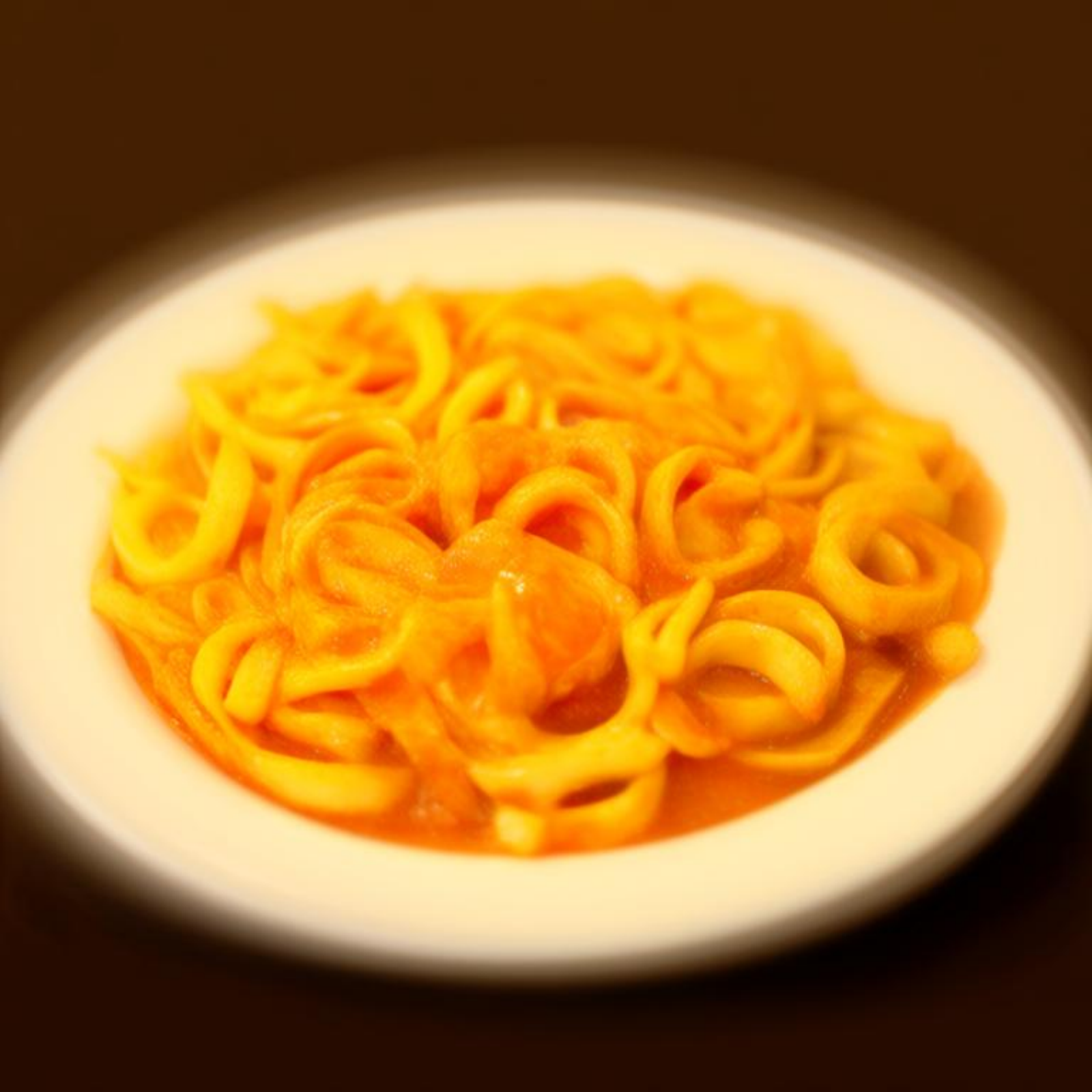}
        \hfill
        \includegraphics[width=0.24\linewidth]{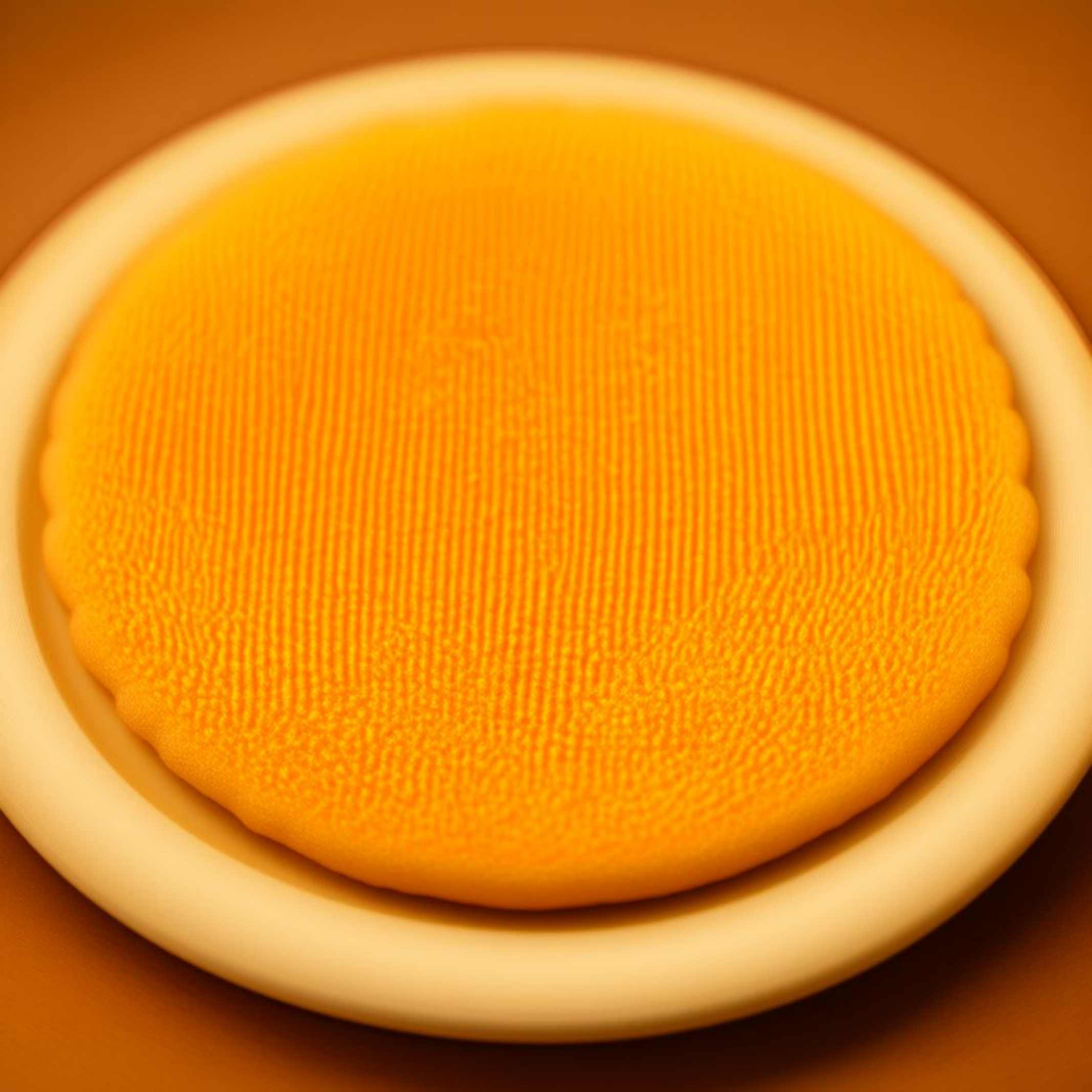}
        \caption{Breakfast: Original CFG.}
    \end{subfigure}
    \vspace{2pt}

    \begin{subfigure}{\linewidth}
        \centering
        \includegraphics[width=0.24\linewidth]{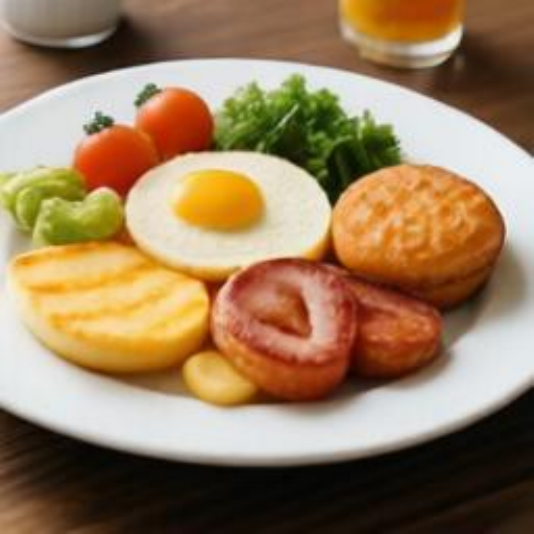}
        \hfill
        \includegraphics[width=0.24\linewidth]{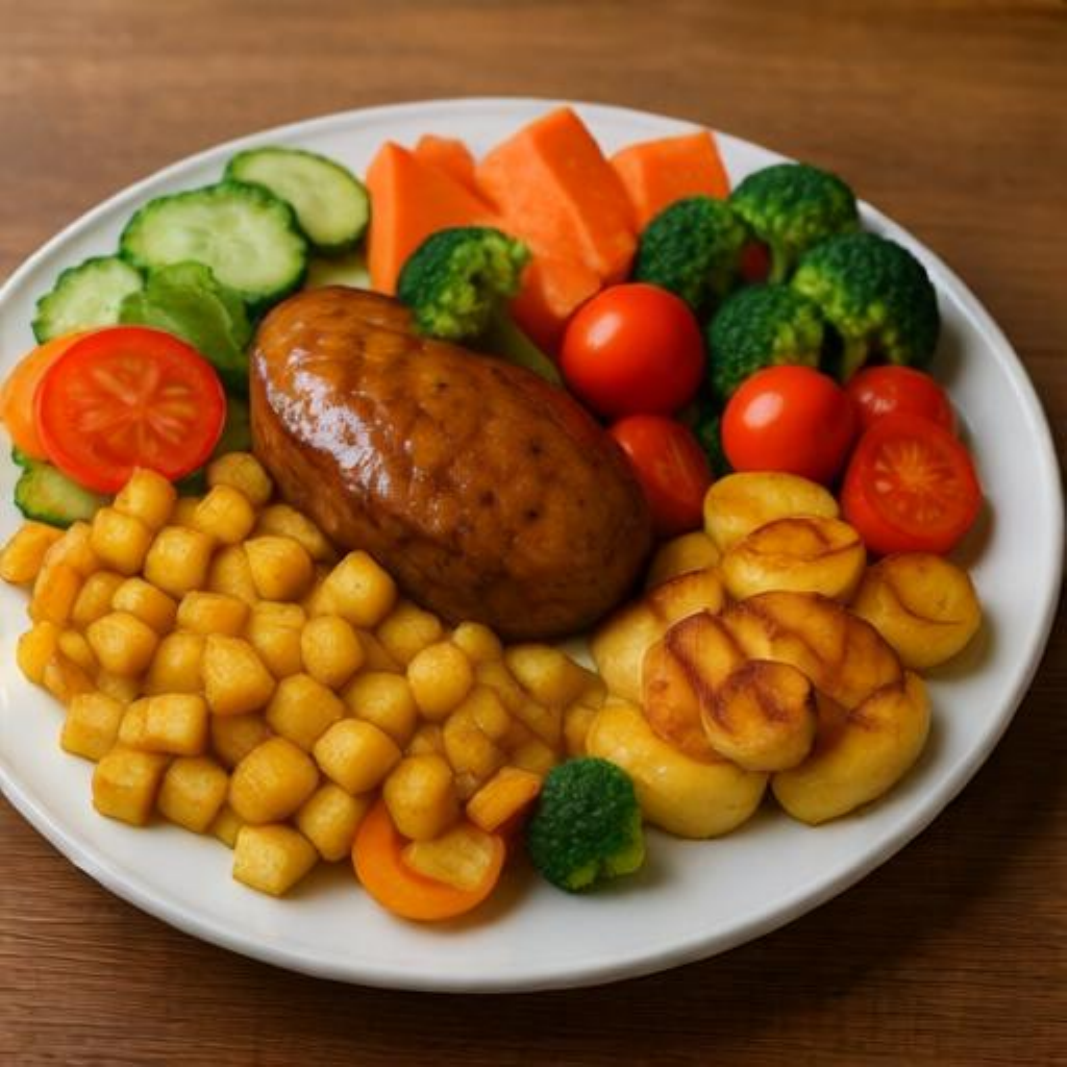}
        \hfill
        \includegraphics[width=0.24\linewidth]{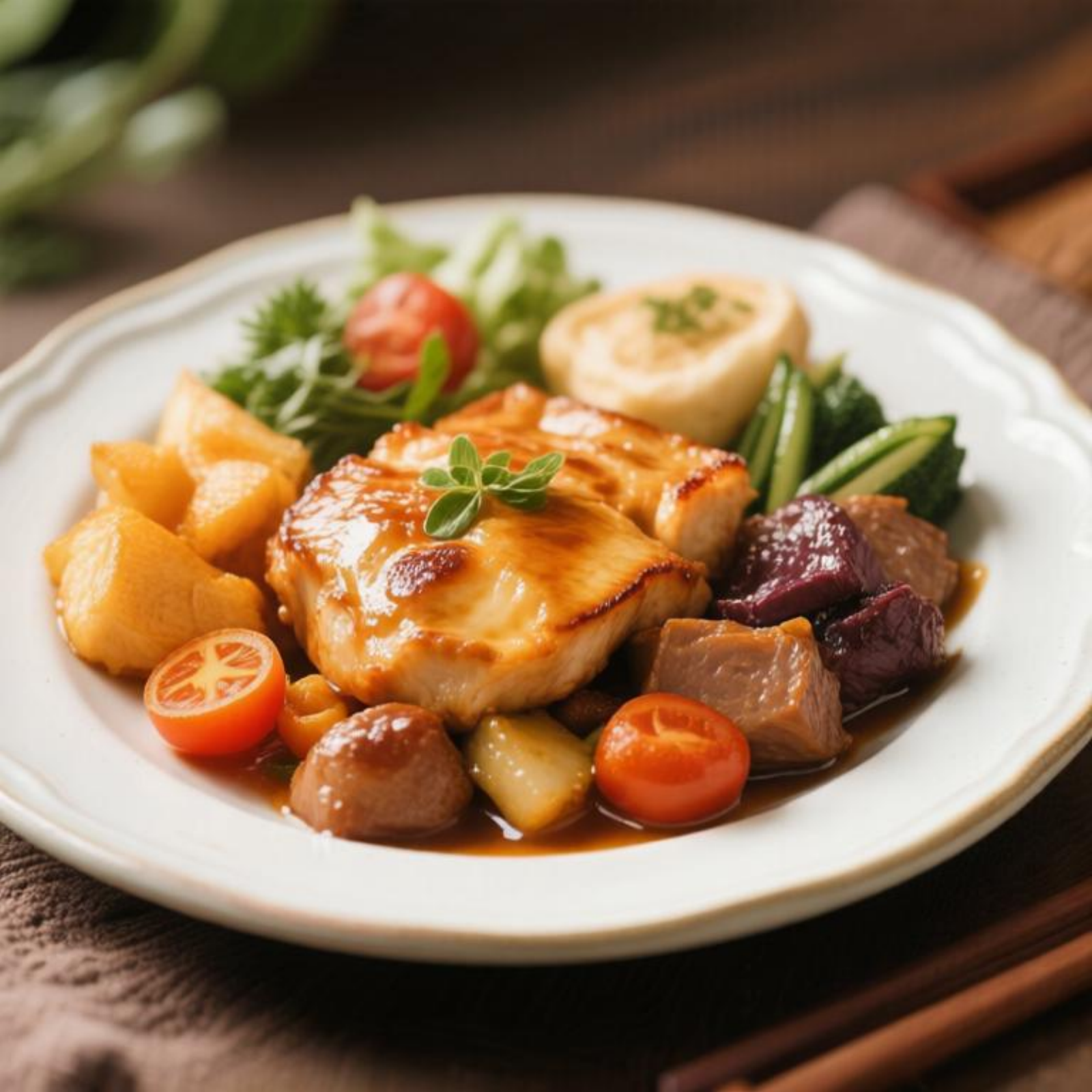}
        \hfill
        \includegraphics[width=0.24\linewidth]{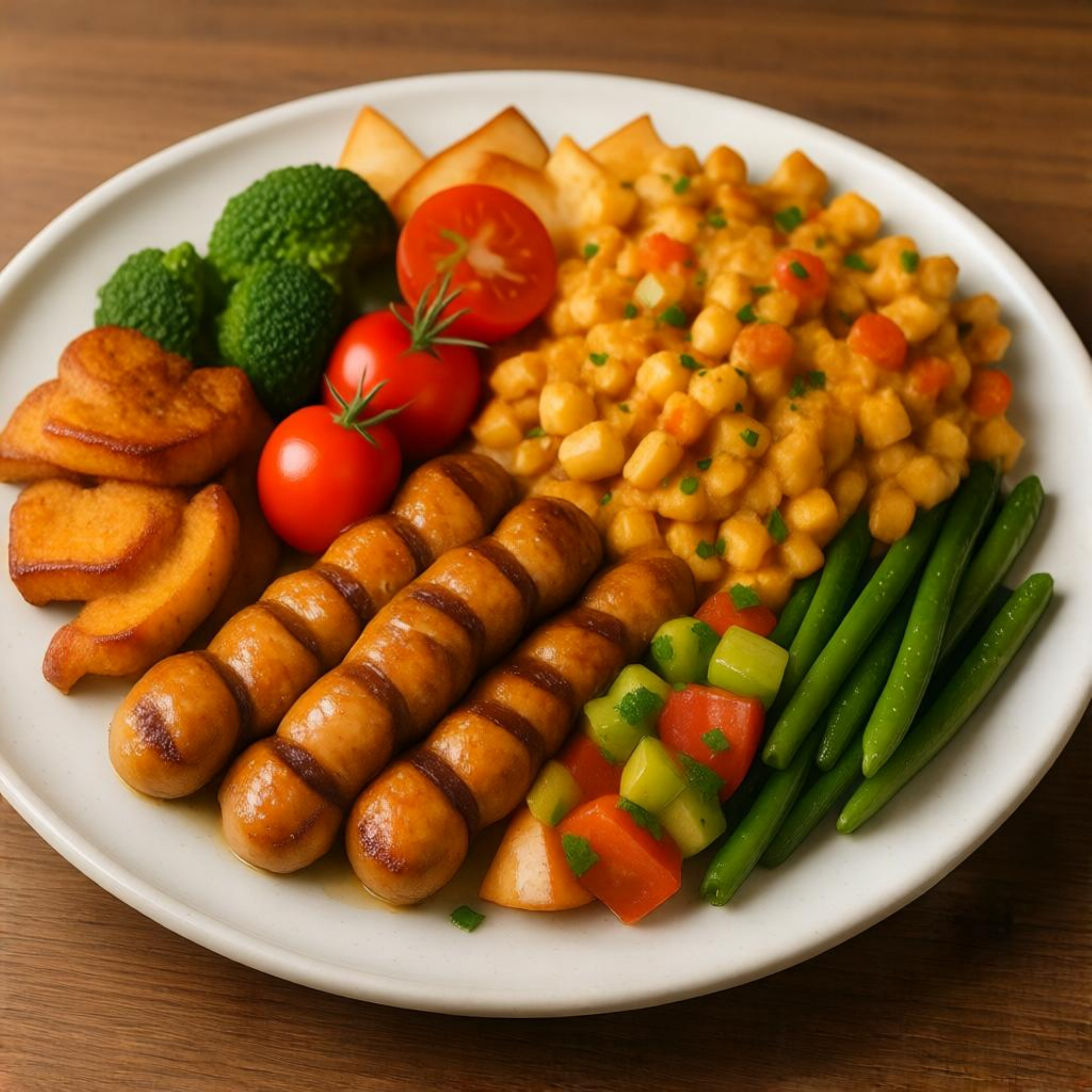}
        \caption{Breakfast: \mtdbk.}
    \end{subfigure}
    
    \caption{\textbf{Visual comparison across resolutions}: These are Qwen-Image~\citep{wu2025qwen} 10-step samples. From the results, it can be seen that with the original CFG, the higher the resolution, the more unstable the model's sampling results are. With \mtdbk, our method stabilizes the sampling process by adjusting the guidance scale at the instance-level, showing significant improvements.}
    \label{fig:resolution_comparison_4rows}
\end{figure*}

\section{Related Work}

\subsection{Guidance Strategies in Diffusion Models}

Various methods aim to improve guidance in diffusion models. \textbf{CFG++}~\citep{chungcfg++} treats guidance as a manifold-constrained inverse problem~\citep{karczewski2025spacetime}, while \textbf{CFG Schedulers}~\citep{xianalysis} and \textbf{Apply Guidance in Interval}~\citep{kynkaanniemi2024applying} optimize time-dependent strength; similarly, \textbf{Stage-Wise Dynamics}~\citep{jin2025stage} investigates varying guidance requirements across different generation stages. \textbf{ReCFG}~\citep{xia2025rectified} and \textbf{TFG}~\citep{ye2024tfg} focus on correcting expectation shifts, a goal shared by \textbf{Rectified CFG++}~\citep{saini2025rectified} and \textbf{CFG-EC}~\citep{yang2025cfg} which propose mechanisms to rectify guidance errors and improve consistency. Recent approaches explore dynamic adaptation: $\mathbf{S^2}$\textbf{-Guidance}~\citep{chen2025s} uses stochastic block-dropping, \textbf{REG}~\citep{gao2025reg} optimizes scaled joint distributions, \textbf{FBG}~\citep{koulischer2025feedback} uses feedback on conditional informativeness, and \textbf{Foresight Guidance}~\citep{wang2025foresight} frames CFG as a fixed-point iteration with short inner loops. Complementary to these, \textbf{Prompt-Aware Guidance}~\citep{zhang2025prompt} adapts strength based on prompt complexity, \textbf{Learning-to-Guide}~\citep{galashov2025learn} employs meta-learning for optimal strategies, and \textbf{Saddle-Free Guidance}~\citep{yeats2025saddle} navigates the optimization landscape to avoid saddle points.
\textbf{Density Guidance}~\citep{zhao2025devil} extends flow matching by incorporating explicit log-density control to steer generation trajectories. While it offers a rigorous theoretical unification of prior heuristics via Score Alignment, the method is computationally expensive. Specifically, estimating the divergence requires Jacobian-Vector Products (JVP), which doubles the inference cost and introduces estimation variance.
In Flow Matching models~\citep{esser2024scalingrectifiedflowtransformers, lipman2022flow, fan2025vchitect, liu2022flow, gao2024lumina}, \textbf{CFG-Zero*}~\citep{fan2025cfg0} compensates for velocity errors.

\subsection{Challenges of Zero-SNR Sampling.} 
\citet{lin2024common} observe that guidance becomes unstable when sampling from true zero-SNR, attributing this to excessive update magnitudes. While Rectified Flow models~\citep{liu2022flow} resolve the training-inference mismatch of standard diffusion schedules by explicitly enforcing a pure Gaussian boundary, they remain susceptible to the zero-SNR guidance instability identified by \citet{lin2024common}. In this work, we analyze why early guidance update leads to instability. Based on this analysis, we propose an adaptive magnitude control mechanism to effectively stabilize the guidance trajectory.

\subsection{Challenges in Large-Scale Generative Models}

Early diffusion models, such as Stable Diffusion v2~\citep{rombach2022high}, operated on relatively compact latent spaces ($\approx 1.6 \times 10^4$ dimensions). In these settings, standard guidance strategies proved robust and forgiving to hyperparameter choices.

However, the transition to modern DiT-based architectures involves a massive increase in scale. For example, Flux~\citep{flux-2-2025} (generating 2K resolution) involves $\approx 10^6$ dimensions, and video models like Wan2.1 (14B)~\citep{wan2.1} exceed $10^7$ dimensions. \textbf{Empirically}, we observe that guidance strategies designed for smaller models become unstable at this scale. 
Building on the observation by \citet{lin2024common} regarding excessive guidance at Zero-SNR, we observe that this phenomenon is further exacerbated by model scale; specifically, the guidance update magnitude scales aggressively with the latent dimensionality.
Without careful regulation, these disproportionately large updates during the initial sampling steps lead to severe visual artifacts, such as color saturation and structural incoherence, effectively causing the model to ``overshoot'' the realistic image distribution. This necessitates a scalable, magnitude-aware correction mechanism.

\section{MAMBO-G: Magnitude-Aware Mitigation}
\subsection{The Risk of Zero-SNR Guidance}
Rectified Flow \cite{liu2022flow} typically initializes sampling from a pure noise state $\x_1\sim\mathcal{N}(\Zero, \One)$ at time $t=1$ to get an example $x_0$ from the original data distribution $\Data$. The velocity field guides the noise $\x_1$ towards the data $\x_0$.
\begin{equation}
    \vel(\x_t, t, c) = \E_{\x_0\sim \Data, \x_1\sim\mathcal N(\Zero, \One)}\left[\left. \x_0 - \x_1 \,\right|\, \x_t, c \right],
\end{equation}
where $\x_t$ is the intermediate state. We denote the conditional velocity as $\vel(\x_t, t, c)$ and the unconditional one as $\vel(\x_t, t, \varnothing)$. The classifier-free guidance substitutes $\vel(\x_t, t, c)$ with $\tilde\vel(\x_t, t, c)$ during sampling, which is defined as:
\begin{equation}
\begin{aligned}
    \tilde\vel(\x_t, t, c) &\\:= w&\cdot(\vel(\x_t, t, c) - \vel(\x_t,t,\varnothing)) + \vel(\x_t,t,\varnothing),
\end{aligned}
\end{equation}
where $w$ is the guidance scale and the \textbf{guidance update} $\Delta\vel(x_t,t,c):=\vel(\x_t, t, c) - \vel(\x_t,t,\varnothing)$.

At the initialization step ($t=1$), the input $\x_1$ is pure noise and statistically independent of the data $\x_0$. Consequently, $\x_1$ provides no spatial or semantic cues about the target image. The model prediction for $\hat\x_0=\E\left[\x_0|\x_t,c\right]$ thus reduces to the conditional expectation that only depends on the text prompt $c$:
\begin{equation}
\begin{aligned}
    \mean_{\cond}&:= \E\left[\left.\x_0 \,\right|\x_1, ~c~ \right]= \E\left[\left.\x_0 \,\right|~c~ \right], \\
    \mean_{\uncond}&:= \E\left[\left.\x_0 \,\right|\x_1, ~c~ \right]= \E\left[\left.\x_0 \,\right|\, \varnothing~ \right].
\end{aligned}
\end{equation}
At this initial step, the guidance update $\Delta \vel(\x_1,1,c)$ reduces to a constant offset:
\begin{equation}
\begin{aligned}
    \Delta \vel_c &= \vel(\x_1, 1, c) - \vel(\x_1, 1, \varnothing) \\
    &=\E\left[\x_0-\x_1|\x_1,c\right] - \E\left[\x_0-\x_1|\x_1,\varnothing\right]\\
    &=(\mean_{\cond} - \x_1) - (\mean_{\uncond} - \x_1)\\
    &= \mean_{\cond} - \mean_{\uncond}.
\end{aligned}
\end{equation}

\begin{figure}[t]
    \centering
    \includegraphics[width=0.48\textwidth]{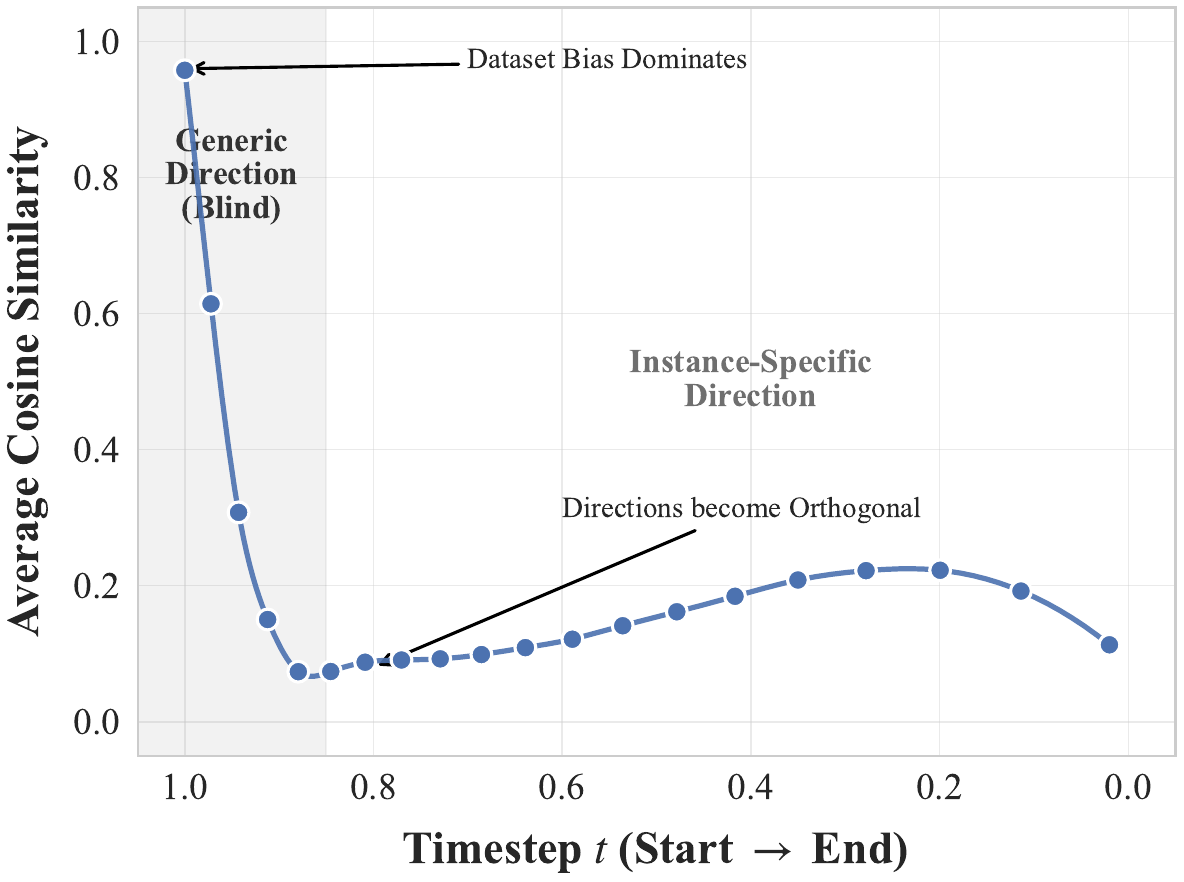}
    \vspace{-0.1in} 
    \caption{\textbf{Collapse of Guidance Directions at Initialization.} We analyze the cosine similarity of guidance updates ($\Delta \vel$) across different noise seeds for a fixed prompt. At $t=1.0$, similarity $\approx 1.0$, indicating a generic direction independent of specific noise. As $t$ decreases ($t < 0.8$), updates rapidly diverge and become instance-specific. This observation motivates \mtdbk to dampen the guidance scale specifically in this high-similarity, generic regime.}
    \label{fig:cosine_sim}
    \vspace{-0.15in}
\end{figure}

\begin{figure}[t]
    \centering
    \includegraphics[width=0.47\textwidth]{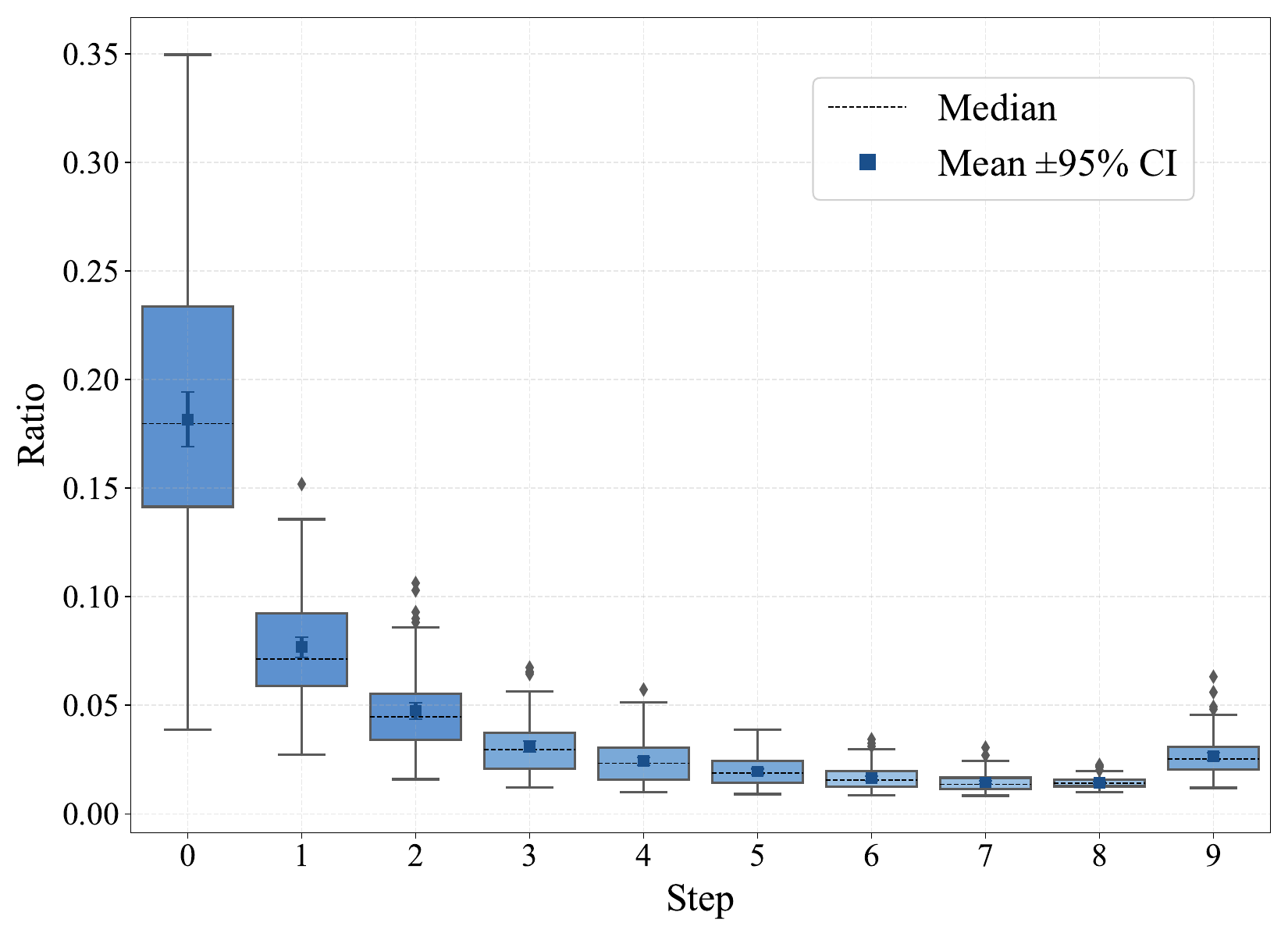} 
    \vspace{-0.1in} 
    \caption{\textbf{Dynamics of the ratio during sampling.} We monitor the evolution of the relative guidance strength $r_t$ throughout the sampling process. The ratio starts at a high peak, reflecting a strong conditional influence that can lead to early-stage instability if left unregulated. It then rapidly decays and stabilizes within a few sampling steps. This empirical trend identifies the initial phase as a critical regime where guidance damping mechanism is most necessary.}
    \label{fig:ratio_distribution_step}
    \vspace{-0.15in}
\end{figure}
This difference $\Delta \vel$ is a \textbf{generic direction} based on dataset statistics. \textbf{It is independent of the sampled noise $\x_1$}.

To understand the guidance behavior at initialization, we analyze the consistency of guidance updates $\Delta \vel$ across different random noise samples for fixed prompts. As shown in \Cref{fig:cosine_sim}, the cosine similarity is approximately \textbf{1.0} at $t=1$. This confirms that the initial guidance is a \textbf{generic direction} determined solely by the prompt, independent of the specific noise $\x_1$.

Applying a large guidance scale to this generic direction is risky. Since the update ignores the specific noise structure, a strong force can drive the trajectory away from the valid data distribution. However, this state is temporary. The similarity drops quickly as sampling continues ($t < 0.8$), and the guidance becomes \textbf{instance-specific}. This motivates \mtdbk: we reduce the guidance scale during this initial generic phase to prevent instability, then boost it to further enhance the guidance effect as the image structure forms.

\subsection{Quantifying the Relative Guidance Strength}
Although the guidance direction $\Delta \vel$ at $t=1$ is independent of $x_1$, the magnitude of the model's conditional velocity $\vel_{\cond}$ is highly instance-dependent. To quantify the relative strength of the guidance update, we define the ratio $r_t$:
\begin{equation}
    r_t = \frac{\norm[2]{\vel(\x_t, t, c) - \vel(\x_t, t, \varnothing)}}{\norm[2]{\vel(\x_t, t, \varnothing)}}.
    \label{eq:ratio}
\end{equation}

From a statistical perspective, $r_t$ provides an \textbf{instance-specific estimation of the relative fluctuation} within the velocity field, functionally analogous to the coefficient of variation (CV). 
This interpretation holds because the unconditional velocity $\vel(\x_t, t, \varnothing)$ approximates the expected trajectory marginalized over the distribution of prompts, while the conditional velocity $\vel(\x_t, t, c)$ acts as a specific realization. 

Intuitively, a high coefficient of variation indicates large relative fluctuations in the guidance-induced discrepancy, reflecting an unstable conditional influence. Such excessive deviation suggests that the conflict between the prompt and the intrinsic image structure is severe, potentially leading to generation artifacts. Therefore, samples with excessively high $r_t$ are risky outliers that require mitigation. 



Empirically, as shown in \Cref{fig:ratio_distribution_step}, this risk is most pronounced during the initialization phase ($t \to 1$), where $r_t$ reaches its peak. This high-ratio phase coincides with the generic direction regime (\Cref{fig:cosine_sim}), where the guidance direction is not yet adapted to the specific noise instance. Consequently, applying a large guidance scale when $r_t$ is high risks amplifying generic, coarse features in an uncontrolled manner, rather than refining the image structure. Thus, $r_t$ serves as a robust indicator for potential instability, necessitating a damping mechanism to prevent trajectory collapse.

\begin{figure}[t]
    \centering
    \includegraphics[width=0.47\textwidth]{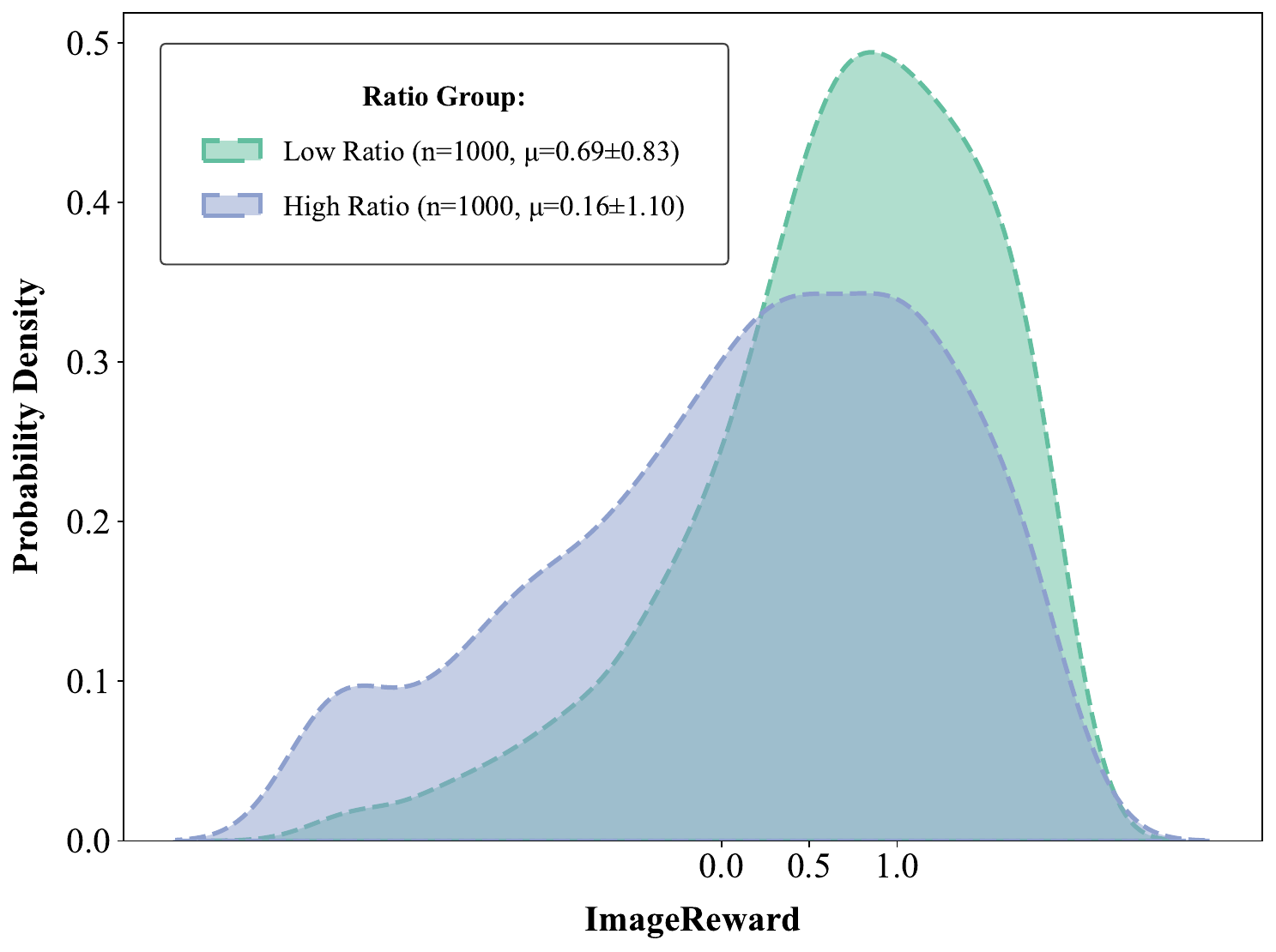} 
    \vspace{-0.1in} 
    \caption{\textbf{Probability density of ImageReward scores across different Ratio groups.} We present KDE plots comparing ImageReward scores for low-ratio versus high-ratio samples at the first sampling step. The results show that lower initial ratios yield significantly higher quality, validating the ratio as a robust indicator for predicting sampling stability.}
    \label{fig:high_ratio_vs_low_ratio}
    \vspace{-0.15in}
\end{figure}

\begin{figure}[t]
    \centering
    \includegraphics[width=0.48\textwidth]{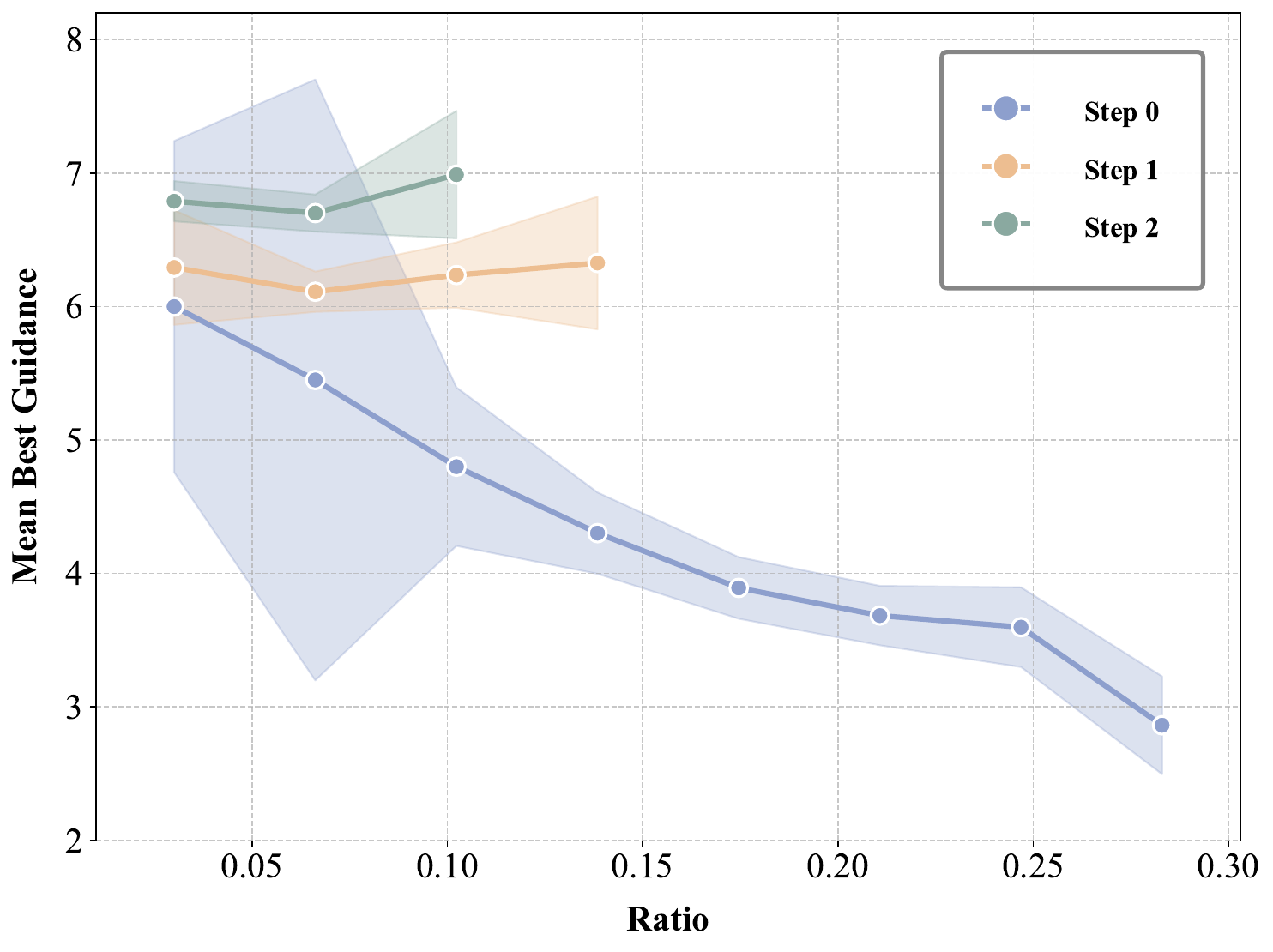}
    \vspace{-0.1in} 
    \caption{\textbf{Optimal Guidance Scale vs. Ratio.} We perform a greedy search to identify the optimal guidance scale maximizing ImageReward for various ratios. The results illustrate that the optimal scale decreases as the ratio increases, exhibiting an exponential decay. This trend directly motivates our use of an exponential damping function.}
    \label{fig:ratio_guidance_fit}
    \vspace{-0.15in}
\end{figure}

\subsection{Empirical Dynamics and Sample Heterogeneity}
Our observations align with \citet{lin2024common} regarding the instability of zero-SNR sampling. To further investigate this, we designed an ablation study on SD3.5. We used 100 prompts, with 20 seeds corresponding to each prompt. The ratio value at the first step was calculated for each of these 20 seeds per prompt, allowing us to divide them equally into high-ratio and low-ratio groups. Sampling was then performed using a default guidance scale of 7. The results, presented in \Cref{fig:high_ratio_vs_low_ratio}, demonstrate that the quality of the high-ratio group was significantly lower than that of the low-ratio group. This experimentally suggests that the ratio can serve as an effective metric for modeling sampling stability. Notably, we observe considerable variance in ratio values across different samples, even at the same timestep $t$. Standard dynamic guidance strategies rely on time-dependent schedules $w(t)$. However, as shown in our analysis, the ratio $r_t$ varies significantly across samples even at the same timestep. A purely time-based schedule $w(t)$ ignores this heterogeneity, failing to selectively mitigate the instability of high-ratio outliers. This observation provides empirical support for our approach of modeling guidance based on the ratio $w(r_t)$ rather than time alone.

\begin{figure*}[t]
    \centering
    \begin{subfigure}{0.49\textwidth}
        \includegraphics[width=\linewidth]{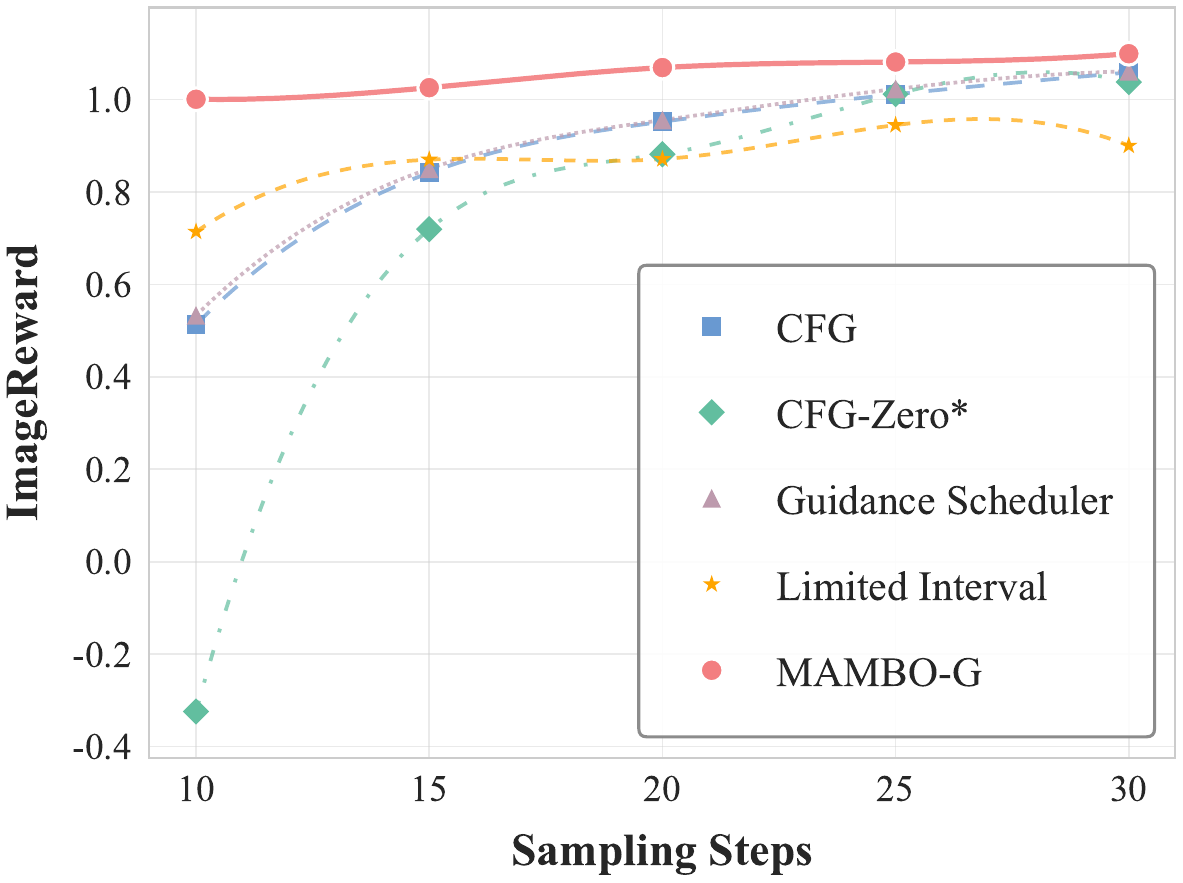}
        \captionsetup{justification=justified, singlelinecheck=true}
        \caption{SD3.5, ImageReward.}
        \label{fig:sub1_1}
    \end{subfigure}
    \hfill
    \begin{subfigure}{0.49\textwidth} 
        \includegraphics[width=\linewidth]{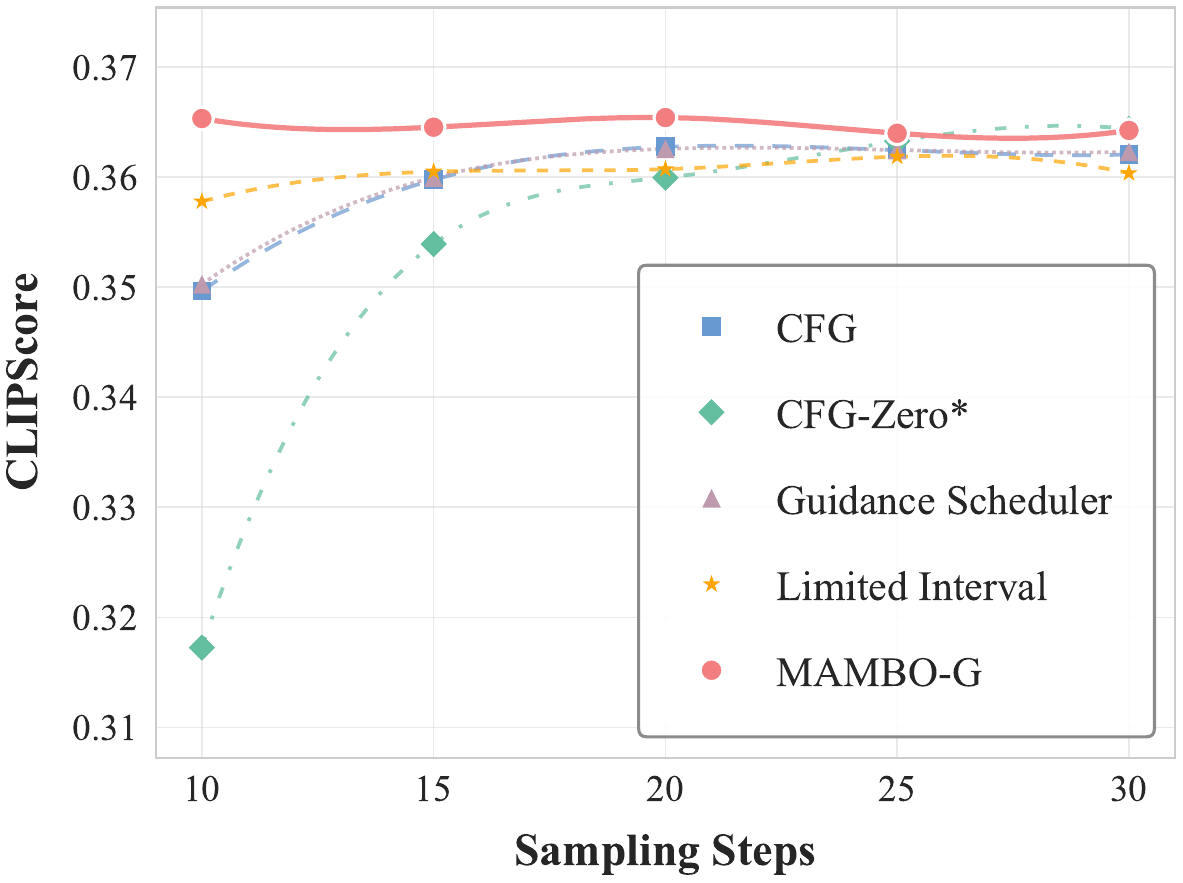}
        \captionsetup{justification=justified, singlelinecheck=true}
        \caption{SD3.5, CLIPScore.}
        \label{fig:sub2_1}
    \end{subfigure}
    \begin{subfigure}{0.49\textwidth}
        \includegraphics[width=\linewidth]{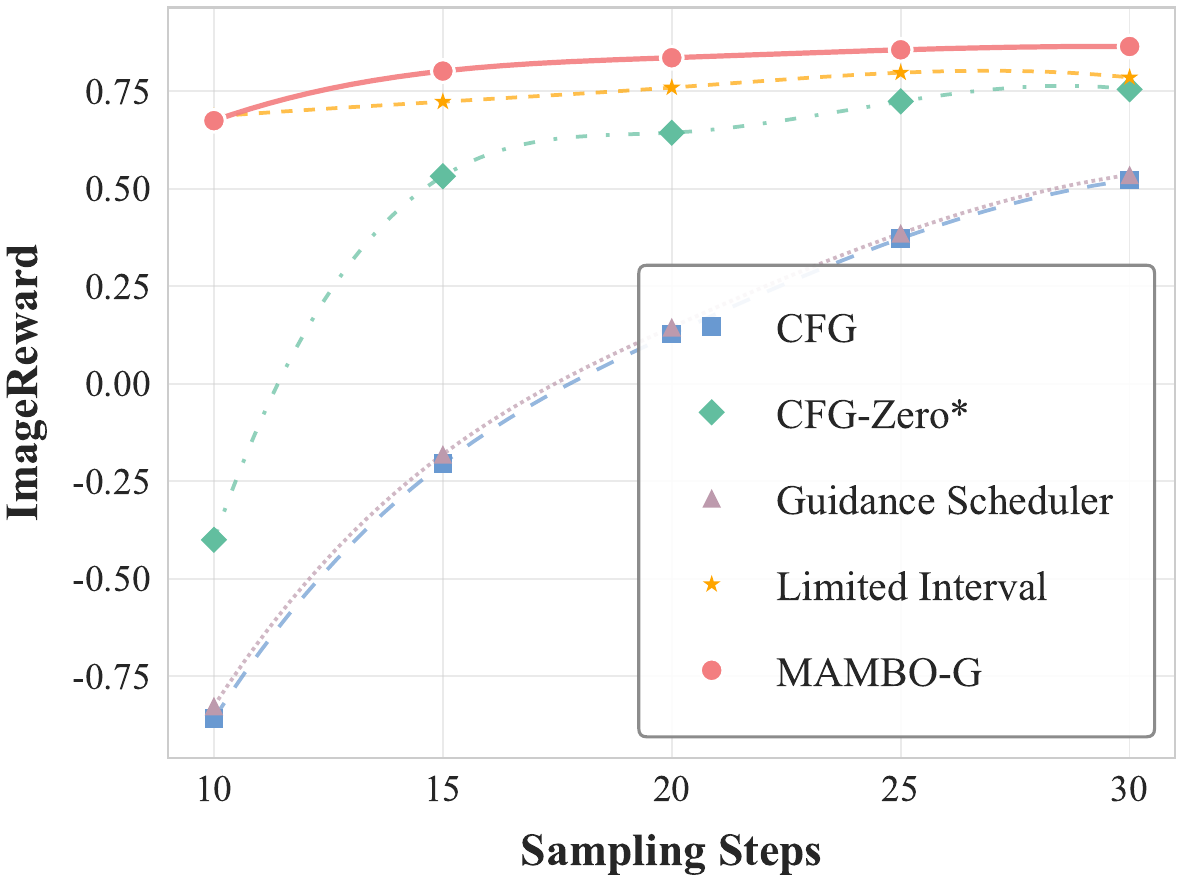}
        \captionsetup{justification=justified, singlelinecheck=true}
        \caption{Lumina, ImageReward.}
        \label{fig:sub1_2}
    \end{subfigure}
    \hfill
    \begin{subfigure}{0.49\textwidth} 
        \includegraphics[width=\linewidth]{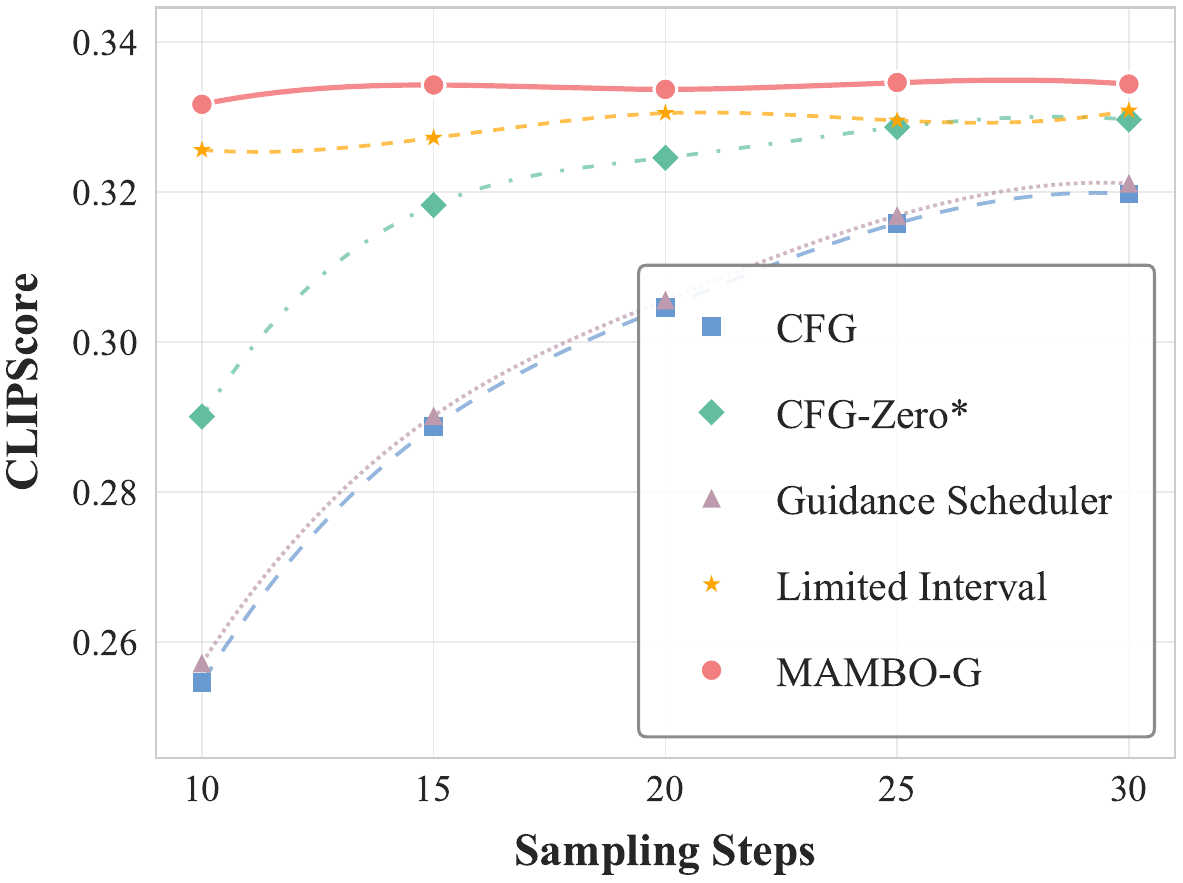}
        \captionsetup{justification=justified, singlelinecheck=true}
        \caption{Lumina, CLIPScore.}
        \label{fig:sub2_2}
    \end{subfigure}
    
    \captionsetup{justification=justified, singlelinecheck=false}
    \caption{\textbf{Quantitative results of comparative analysis with other baselines in text-to-image generation measured by ImageReward and CLIPScore.} Here, \textbf{Guidance Scheduler} refers to ~\citet{xianalysis} and \textbf{Limited Interval} refers to ~\citet{kynkaanniemi2024applying}. The results demonstrate \methodabbr's remarkable superiority over others in low-step generation, achieving the quality of $30$-step generation of CFG in only \textbf{$\mathbf{10}$ steps.}}
    \label{fig:txt2img_compare}
\end{figure*}
\label{sec:exp}

\subsection{MAMBO-G: Adaptive Damping Strategy}

Building on the statistical insight that updates with high $r_t$ represent outliers (Eq.~\ref{eq:ratio}), we formulate the damping function $w(r_t)$ to mitigate the risk of these deviations. Since high-$r_t$ updates likely exceed the valid velocity distribution, relying on them with a strong guidance scale typically leads to trajectory collapse or artifacts.

To determine the optimal suppression schedule, we turn to the empirical evidence. By performing a controlled grid search for the maximum effective guidance scale across varying $r_t$ (see \Cref{alg:greedy_search}), we derive an empirical reference curve (visualized in \Cref{fig:ratio_guidance_fit}). We observe that the model's tolerance for strong guidance does not decay linearly, but drops sharply as the update becomes more disproportionate. To strictly align the guidance strength with this safe regime, we fit the stability boundary with an exponential decay function:
\begin{equation}
    w(r_t) = 1 + (w_{\max} - 1) \cdot \exp(-\alpha r_t).
\end{equation}
Here, $w_{\max}$ represents the maximum allowable guidance scale, and $\alpha > 0$ is a hyperparameter calibrated to the decay rate of the reference curve.

This formulation acts as a continuous, magnitude-aware filter. It permits aggressive boosting when the conditional update is statistically normal ($r_t \to 0$) but applies exponentially stronger damping as $r_t$ increases. By dynamically suppressing the specific "outlier" updates identified by $r_t$, \mtdbk prevents the amplification of unstable directions while retaining the benefits of high guidance in safe regions.

\begin{figure*}[t]
    \centering
    \begin{subfigure}{0.49\textwidth}
        \includegraphics[width=\linewidth]{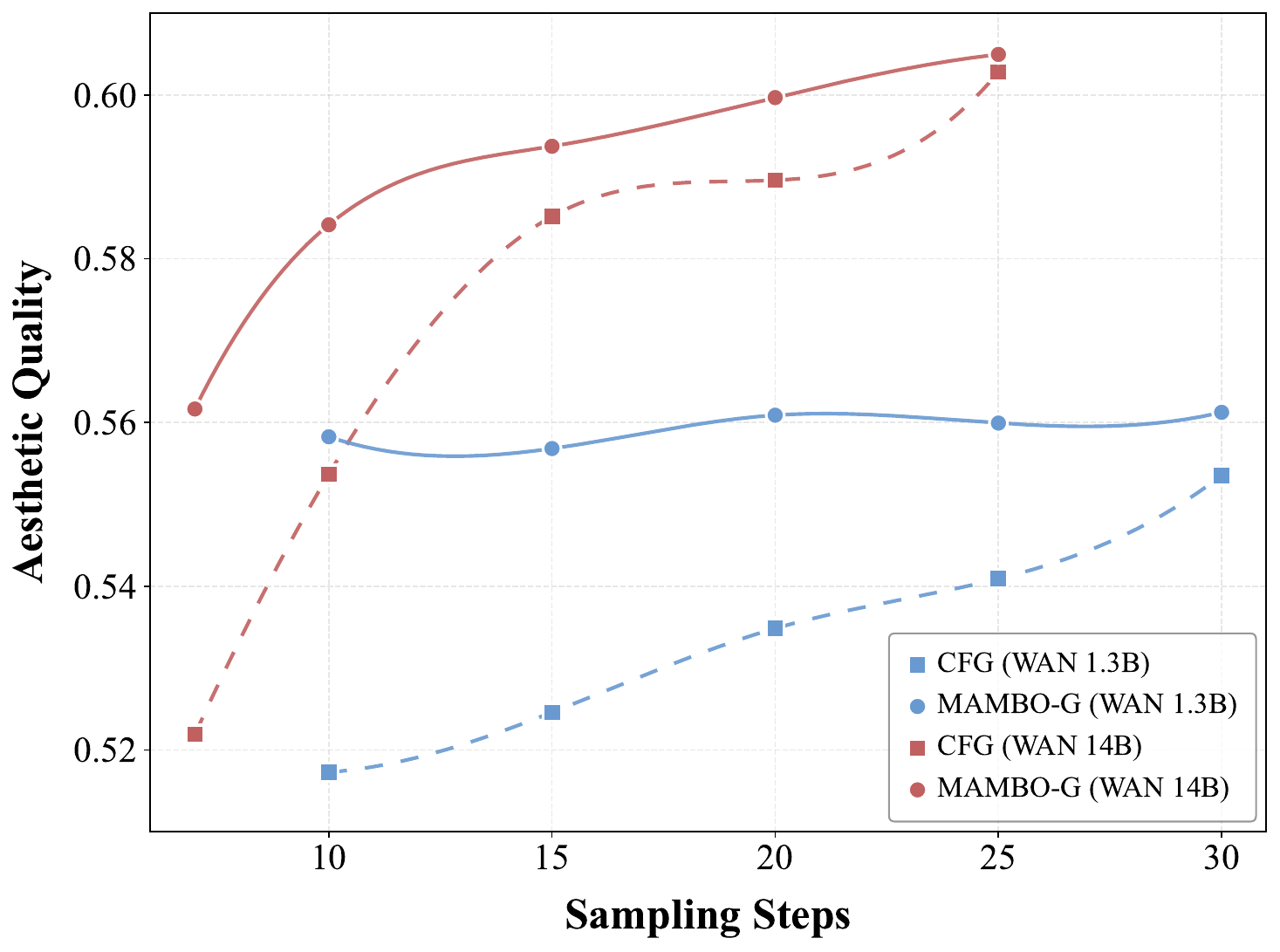}
        \captionsetup{justification=justified, singlelinecheck=true}
        \caption{vBench: Imaging Quality.}
        \label{fig:sub1_1}
    \end{subfigure}
    \hfill
    \begin{subfigure}{0.49\textwidth} 
        \includegraphics[width=\linewidth]{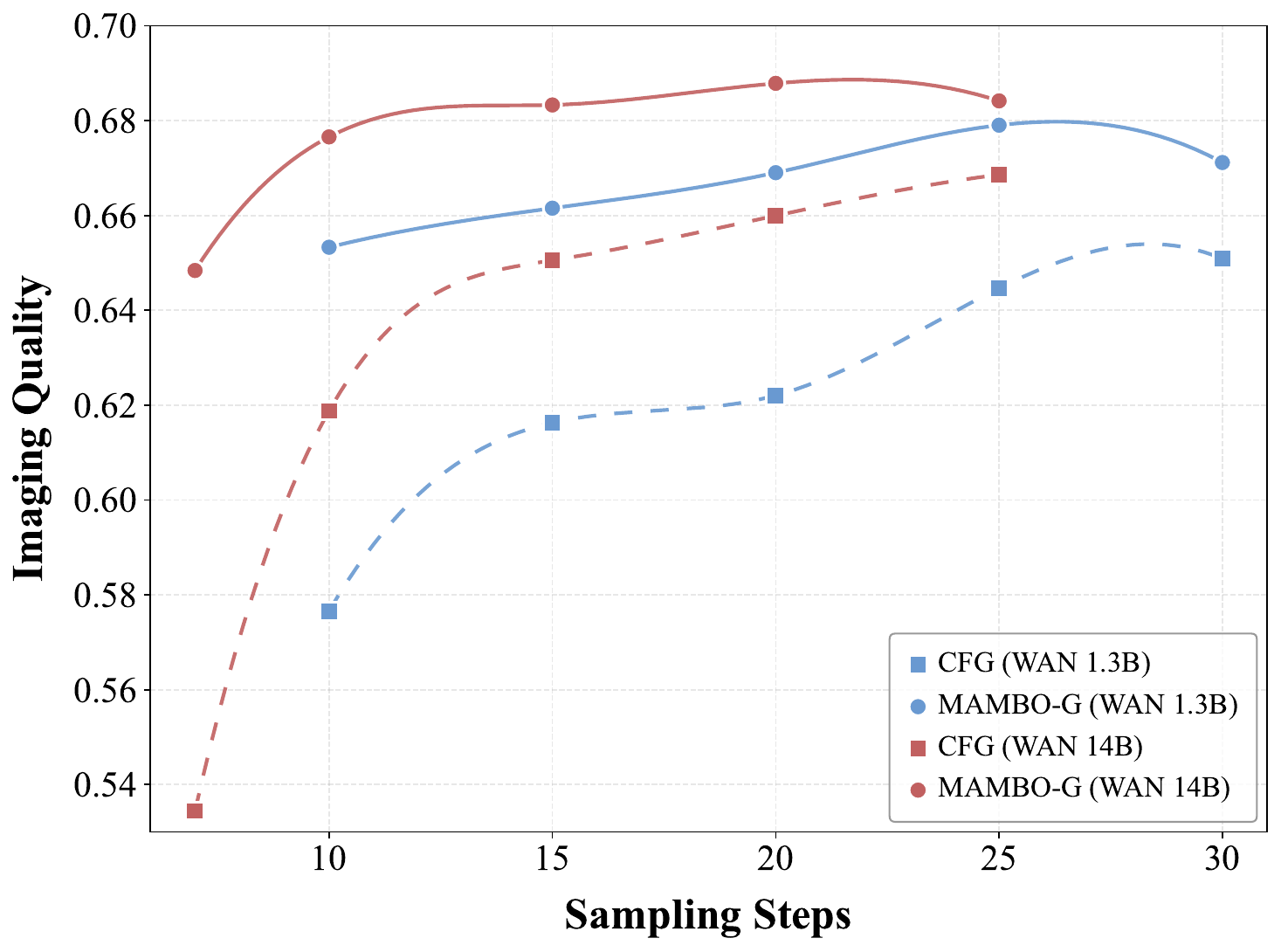}
        \captionsetup{justification=justified, singlelinecheck=true}
        \caption{vBench: Aesthetic Quality.}
        \label{fig:sub2_1}
    \end{subfigure}
    \captionsetup{justification=justified, singlelinecheck=false}
    \caption{\textbf{Quantitative results of text-to-video generation, comparing \textbf{CFG and \mtdbk} under vBench.} The results strongly validate the effectiveness of \mtdbk on video generations, even overtaking CFG-Wan 14B just on Wan 1.3B.}
    \label{fig:txt2video_compare}
\end{figure*}
\label{sec:exp}
\section{Experiments}
\begin{figure*}[hbt]
    \centering
    \begin{subfigure}{0.49\textwidth}
        \includegraphics[width=\linewidth]{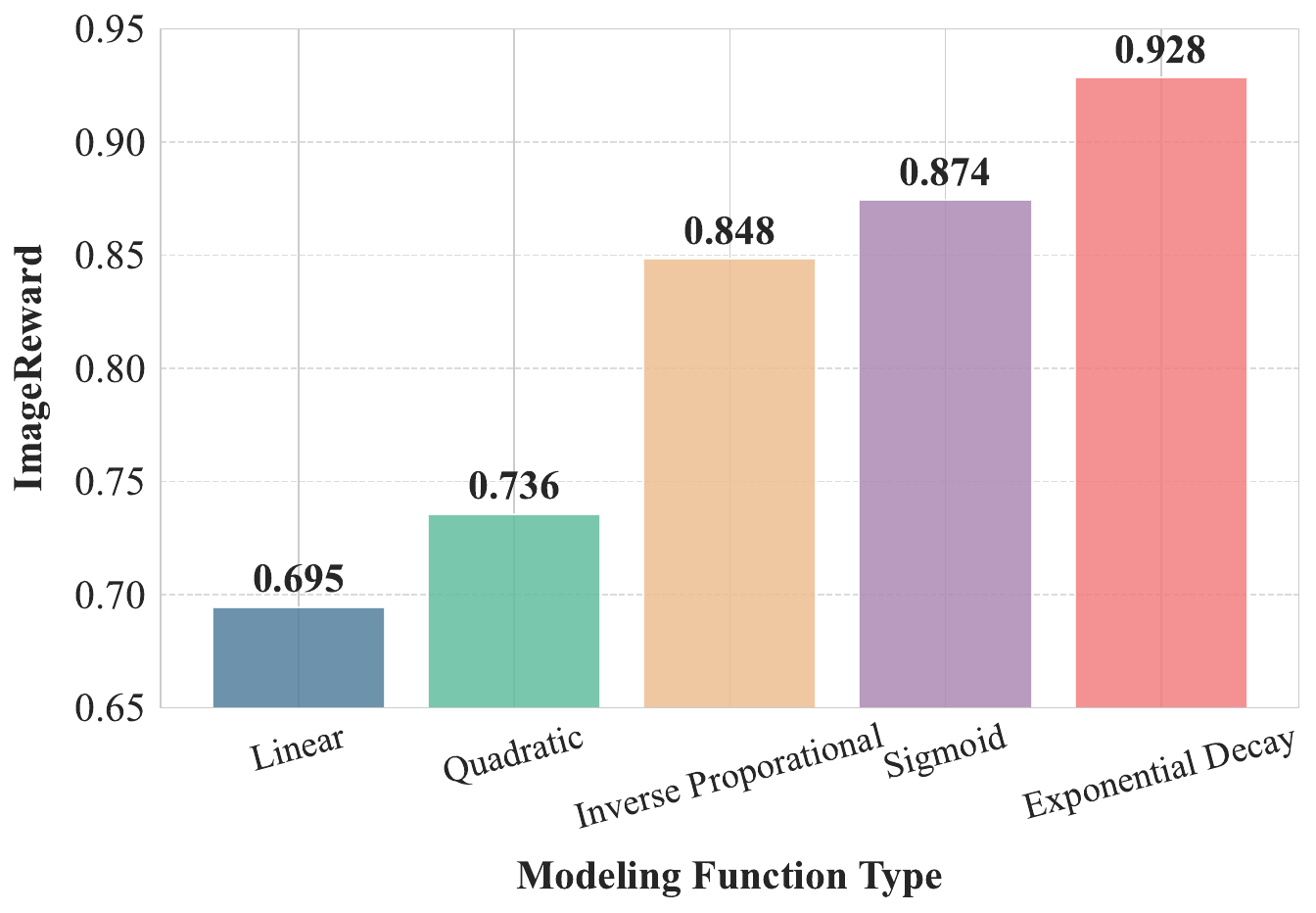}
        \captionsetup{justification=justified, singlelinecheck=true}
        \caption{Modeling function comparison: ImageReward.}
        \label{fig:sub1}
    \end{subfigure}
    \hfill
    \begin{subfigure}{0.49\textwidth}
        \includegraphics[width=\linewidth]{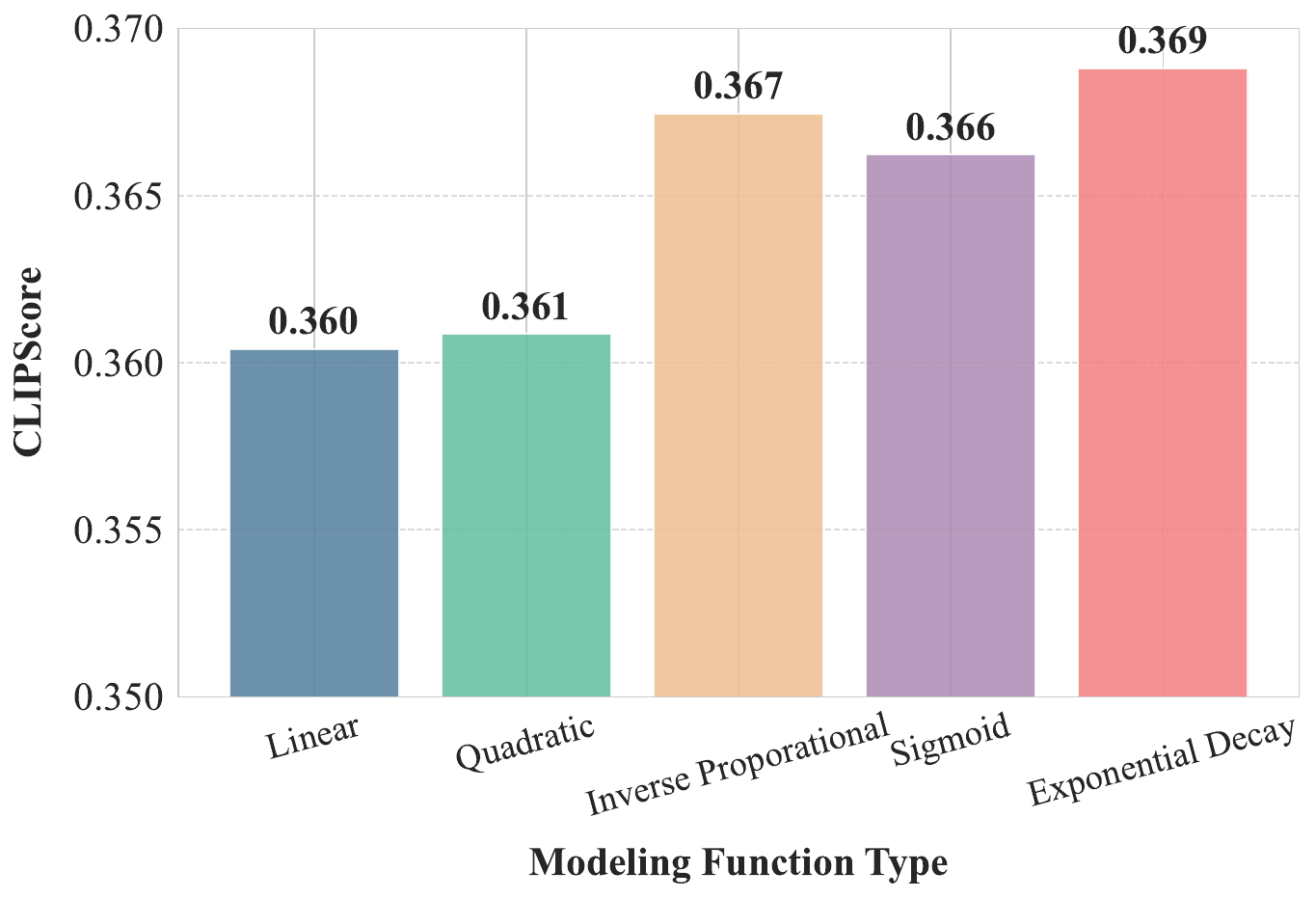}
        \captionsetup{justification=justified, singlelinecheck=true}
        \caption{Modeling function comparison: CLIPScore.}
        \label{fig:sub2}
    \end{subfigure}
    \captionsetup{justification=justified, singlelinecheck=false}
    \caption{\textbf{Ablation studies on different types of modeling functions.} The comparative results demonstrate that the exponential decay function delivers the optimal fitting results among other functions, validating the soundness of our method.}
    \label{fig:function_ablation}
\end{figure*}
We evaluate \mtdbk on image and video generation, testing its effectiveness across architectures. We also study how design choices and hyperparameters affect performance.
\subsection{Text-to-Image Generation}
\label{subsec:txt2img_generation}

We test \mtdbk on text-to-image generation using two recent models: Stable Diffusion v3.5 (SD3.5) and Lumina-Next (see \Cref{fig:txt2img_compare}). We measure quality with ImageReward and CLIPScore, comparing against the base samplers.
Across both models, \mtdbk improves image quality and semantic alignment. With fewer sampling steps, \mtdbk matches or exceeds longer-step baselines, speeding up generation without changing the underlying sampler.
We measure its improvement by FID as well (see \Cref{app:exp_fid}).

\subsection{Text-to-Video Generation}

We apply \mtdbk to video diffusion models and evaluate on vBench metrics for visual quality and aesthetics (see \Cref{fig:txt2video_compare}). Compared with the original guidance schedules, models with \mtdbk produce videos with higher quality and better semantic alignment at the same step count. In some cases, a smaller backbone with \mtdbk matches the quality of a larger baseline.

\subsection{Impact of Dimensionality}
\label{subsec:dim_analysis}


\begin{table}[h]
    \centering
    \caption{\textbf{Quantitative results of text-to-image generation across different resolutions measured by \textbf{ImageReward}.} \mtdbk demonstrates consistent performance, whereas CFG degrades significantly at higher resolutions.}
    \label{tab:resolution_ir}
    \small 
    \begin{tabular}{ccc}
        \toprule
        Resolution & CFG (Baseline) & \textbf{\mtdbk (Ours)} \\
        \midrule
        $256 \times 256$   & 0.53 & \textbf{0.83} \\
        $512 \times 512$   & 0.63 & \textbf{1.10} \\
        $768 \times 768$   & 0.30 & \textbf{1.07} \\
        $1024 \times 1024$ & 0.20 & \textbf{1.02} \\
        \bottomrule
    \end{tabular}
\end{table}

To verify that guidance instability increases with dimensionality, we evaluate \mtdbk across different resolutions using Qwen-Image, ranging from $256\times256$ to $1024\times1024$.

We compare the ImageReward of the baseline guidance against \mtdbk. As shown in \Cref{tab:resolution_ir}, the performance gap widens significantly as resolution increases. At lower resolutions, the baseline performs adequately, and the gain from \mtdbk is marginal. However, at $1024\times1024$, the baseline frequently suffers from over-saturation and artifacts, whereas \mtdbk maintains structural coherence (see \Cref{fig:resolution_comparison_4rows}). This trend confirms that the risk of unscaled guidance updates is inherently linked to the total noise magnitude in high-dimensional spaces, making our method particularly vital for future high-resolution video models.

\subsection{Compatibility with Advanced Guidance Strategies}
\label{subsec:orthogonality}
\begin{table}[h]
    \centering
    \caption{\textbf{Orthogonality analysis of \mtdbk with other methods.} The results show that \mtdbk can be seamlessly integrated with other methods like APG \cite{sadat2024eliminating} and Rescale \cite{lin2024common} to further improve performance.}
    \label{tab:orthogonality}
    \begin{tabular}{lc}
        \toprule
        Method & ImageReward \\
        \midrule
        Baseline (Constant CFG) & 0.12 \\
        \midrule
        Rescale & 0.73 \\
        Rescale + \mtdbk & \textbf{1.12} \\
        \midrule
        APG & 0.85 \\
        APG + \mtdbk & \textbf{0.96} \\
        \bottomrule
    \end{tabular}
\end{table}
A key advantage of \mtdbk is its orthogonality to other guidance optimization
techniques. Since our method exclusively modulates the guidance scale, it can be directly stacked with methods that normalize the update vector or alter the sampling trajectory.

To verify this, we evaluate \mtdbk in combination with Guidance Rescale (GR)~\cite{lin2024common} and Adaptive Projection Guidance (APG) using the Qwen-Image model. GR rescales the guidance vector to prevent over-exposure, while APG dynamically adjusts the projection direction. As shown in our comparisons (\Cref{tab:orthogonality}), while both baselines effectively mitigate some artifacts, stacking them with \mtdbk yields further consistent improvements in ImageReward. This plug-and-play nature ensures \mtdbk remains relevant even as new guidance strategies are developed.

\subsection{Ablation Studies and Hyperparameters}
\label{subsec:ablation}

We perform ablation studies to understand the impact of our modeling choices.

\textbf{Guidance schedule.} We compare several ways of mapping $r_t$ to a guidance scale, including exponential decay, linear decay, and simple inverse functions. Exponential decay provides a good balance between stability and detail in our experiments (see \Cref{fig:function_ablation}), so we use it as the default.

\begin{table}[h]
    \centering
    \caption{\textbf{Comparison with time-based schedule on Qwen-Image (10 steps).} The time-based schedule uses the average guidance scale of \mtdbk at each step. The results highlight the importance of instance-level adaptation.}
    \label{tab:time_vs_instance}
    \begin{tabular}{lc}
        \toprule
        Method & ImageReward \\
        \midrule
        Baseline (Constant CFG) & 0.12 \\
        Time-based Schedule & 0.83 \\
        \mtdbk (Ours) & \textbf{1.08} \\
        \bottomrule
    \end{tabular}
\end{table}

\textbf{Instance-aware vs. Time-based Schedule.} A natural question is whether a simple time-dependent schedule suffices. To investigate this, we constructed a ``Time-based Schedule'' baseline by averaging the effective guidance scale $w(r_t)$ of \mtdbk across all samples at each timestep, then applying this fixed curve to all generated images. As shown in \Cref{tab:time_vs_instance}, the time-based schedule significantly improves over the constant baseline (ImageReward $0.12 \to 0.83$), confirming that dampening early-stage guidance is generally beneficial. However, \mtdbk achieves a further substantial improvement ($0.83 \to 1.08$). This gap underscores the critical role of \textit{instance-awareness}: since instability varies across random seeds, a fixed schedule over-penalizes stable samples or under-penalizes risky ones, whereas \mtdbk adapts dynamically.

\textbf{Hyperparameter sensitivity.} We vary $w_{\max}$ and the decay rate $\alpha$ over a range of values (see \Cref{tab:wmax_ablation,tab:alpha_ablation}). \mtdbk maintains stable behavior and competitive scores across a broad region, suggesting it does not require heavy tuning.

\textbf{Scheduler generalization.} We also apply \mtdbk on ODE solvers such as UniPC (see \Cref{fig:unipc_ablation}). The method still improves over the corresponding baselines, showing it can be combined with different samplers.


\section{Conclusion}

In this work, we address the instability of Classifier-Free Guidance (CFG) in large-scale models by identifying risks from excessive magnitudes during initialization. We propose \mtdbk, a training-free strategy that dynamically modulates the guidance scale based on the update-to-base ratio $r_t$. Unlike fixed scales, \mtdbk adaptively prevents artifacts during critical early steps, ensuring realistic generation trajectories. Experiments on text-to-image and text-to-video tasks demonstrate that \mtdbk significantly accelerates inference and stabilizes high-resolution generation without architectural changes. Our results highlight magnitude-aware control as a robust, efficient component for state-of-the-art foundation models.

\bibliography{example_paper}
\bibliographystyle{icml2026}

\newpage
\appendix
\onecolumn
\section{Experimental Settings}

In this section, we provide a comprehensive description of the experimental configurations to ensure the reproducibility of our results. We detail the model specifications, sampling schedulers, datasets, and randomization protocols employed in our evaluations.

\subsection{Models and Schedulers}
\label{app:models_schedulers}

We conduct experiments across a diverse set of generative models. The specific sampling schedulers and hyperparameter configurations for each model are detailed below:

\begin{itemize}
    \item \textbf{Stable Diffusion v3.5}:
    We employ the Flow Matching Euler scheduler by default, while the UniPC scheduler is employed for ablation studies. For the Classifier-Free Guidance (CFG) baseline, a guidance scale of $7.0$ is applied.

    \item \textbf{Lumina-Next}:
    Inference is performed using the UniPC scheduler. The guidance scale is set to $7.0$ for the CFG baseline.

    \item \textbf{Wan2.1}:
    Similarly, this model uses the UniPC scheduler; however, the guidance scale is adjusted to $5.0$ for the CFG baseline to ensure optimal performance in video generation.

    \item \textbf{Wan2.2}: This model uses Flow Matching Euler scheduler; the guidance scale is adjusted to $5.0$ for the CFG baseline to ensure optimal performance in video generation.

    \item \textbf{Qwen-Image}: This model uses Flow Matching Euler scheduler; the guidance scale is adjusted to $4.0$ for the CFG baseline to ensure optimal performance in video generation.
    
\end{itemize}

\subsection{Datasets and Randomization}
\label{app:datasets}

To guarantee reliability and deterministic generation, we specify the datasets and seed strategies used for quantitative evaluation in the main paper:

\begin{itemize}
    
    \item \textbf{Text-to-Image Generation (\Cref{fig:txt2img_compare})}:
    Evaluations are performed on the \textbf{MS-COCO}. We construct a test set comprising the initial $500$ prompts extracted from the full dataset.
    \begin{itemize}
        \item \textbf{Randomization}: To ensure reproducibility, we adopt a deterministic strategy where the random seed for each sample is set equal to its corresponding prompt index.
    \end{itemize}

    \item \textbf{Text-to-Video Generation (\Cref{fig:txt2video_compare})}:
    Video synthesis capabilities are evaluated using \textbf{WebVid}. For this task, we randomly select a subset of $100$ samples.
    \begin{itemize}
        \item \textbf{Randomization}: The randomization protocol remains consistent with the text-to-image setting (i.e., seed equals sample index).
    \end{itemize}

    \item \textbf{Impact of Dimensionality (\Cref{tab:resolution_ir})}:
    A specific subset consisting of the first 200 prompts from \textbf{ImageReward Dataset~\citep{xu2023imagereward}} is employed for this analysis.
    \begin{itemize}
        \item \textbf{Randomization}: To maintain fixed noise initialization across experiments, the random seed is assigned to match the prompt ID of each sample.
    \end{itemize}

    \item \textbf{Orthogonality Analysis (\Cref{tab:orthogonality})}:
    We use a curated collection of 200 prompts drawn from \textbf{ImageReward Dataset}.
    \begin{itemize}
        \item \textbf{Randomization}: Stochasticity is governed by a deterministic mapping, where the seed for each generation is aligned with the prompt's index in the test suite.
    \end{itemize}

    \item \textbf{Comparison with Time-Based Schedule (\Cref{tab:time_vs_instance})}:
    Testing is conducted using the first 200 samples extracted from \textbf{ImageReward Dataset}.
    \begin{itemize}
        \item \textbf{Randomization}: We adhere to a predefined scheme in which the random seed for each instance is set equal to its sequence number in the group.
    \end{itemize}
    
\end{itemize}

\subsection{Hyper-Parameter Settings}

 For all experiments involving \mtdbk, we use the default parameters $\alpha = 8$ and $w_{\max} = 10$, which provide robust results across diverse scenarios.

In addition to \mtdbk, we detail the configurations for all baselines discussed in \Cref{subsec:txt2img_generation} as follows:

\begin{itemize}
 \item \textbf{CFG-Zero*~\citep{fan2025cfg0}:} This method is parameter-free. We strictly follow the implementation details provided in the original work.

 \item \textbf{Guidance Scheduler~\citep{xianalysis}:} We employ a cosine schedule for guidance and set the minimum guidance scale to $4.0$.

 \item \textbf{Limited Interval~\citep{kynkaanniemi2024applying}:} CFG is applied during the interval spanning $10\%$ to $90\%$ of the total sampling steps (i.e., excluding the first and last $10\%$). The guidance scale is fixed at $7.0$.
 \end{itemize}

\subsection{Prompts for Visual Examples}
\label{app:prompts}

In this subsection, we list the exact textual prompts corresponding to the visual qualitative results presented in the main paper.

\begin{description}
    \item[\Cref{fig:display} (a):]
    \textit{"Cirno, Touhou, ice fairy, light blue short hair, big blue hair ribbon, blue eyes, smug face, open mouth, confidence, blue and white dress, serrated skirt, red neckerchief, crystal wings, ice wings, floating, hands on hips, ice magic, snowflakes, frozen lake background, misty, magical atmosphere, anime style, cel shading, masterpiece, best quality, 8k, vivid colors"}

    \item[\Cref{fig:display} (b):]
    \textit{"A photograph of an astronaut riding a horse, high quality, 4k, detailed, on Moon"}

    \item[\Cref{fig:display} (c) (d):]
    \textit{"Two anthropomorphic cats in comfy boxing gear and bright gloves fight intensely on a spotlighted stage."}

    \item[\Cref{fig:resolution_comparison_4rows}:]
    \textit{"delicious plate of food"}
\end{description}

\subsection{Configurations of Exploratory Experiments}
\label{app:config_explore}

Here we specify the detailed settings for the exploratory analyses discussed in the main text.

\begin{itemize}
    \item \textbf{Collapse of Guidance Directions at Initialization (\Cref{fig:cosine_sim}):}
    These results are derived from Stable Diffusion v3.5 using a default guidance scale of $7.0$. We configure the random noise seeds and use the fixed prompts from MS-COCO.

    \item \textbf{Dynamics of the ratio during sampling (\Cref{fig:ratio_distribution_step}):}
    Ratio values are tracked using Stable Diffusion v3.5 with a guidance scale of $7.0$. The sampling process consists of 10 steps. The dataset comprises $100$ prompts from MS-COCO, following the randomization strategy defined in \Cref{subsec:txt2img_generation}.

    \item \textbf{Probability density of ImageReward scores across different ratio groups (\Cref{fig:high_ratio_vs_low_ratio}):}
    Using Stable Diffusion v3.5 (guidance scale $7.0$), we analyze the first $20$ prompts from MS-COCO. We employ a 10-step sampling procedure for each generation. For each prompt, we generate samples across $20$ distinct seeds. Ratio groups are determined via binary splitting based on the median ratio value.

    \item \textbf{Optimal Guidance Scale vs. Ratio (\Cref{fig:ratio_guidance_fit})}:
    We search for optimal guidance scales relative to observed ratio values on Stable Diffusion v3.5. A constant guidance schedule of $7.0$ serves as the baseline. The generation process uses 10 sampling steps. For the first step, we sweep the guidance scale from $1.5$ to $9.0$ in increments of $0.5$ to identify the optimal configuration, then searching for the subsequent step based on the previously optimal guidance schedule. Detailed algorithm is presented in \Cref{alg:greedy_search}. This experiment uses prompts from MS-COCO, adhering to the randomization strategy described in \Cref{subsec:txt2img_generation}.

\end{itemize}

\begin{algorithm}[t]
\caption{Greedy Search for Step-wise Optimal Guidance Scale}
\label{alg:greedy_search}
\begin{algorithmic}[1]
\REQUIRE Prompts $\mathcal{P}$ from MS-COCO, Sampling steps $T=10$, Search space $\mathcal{W} = \{1.5, 2.0, \dots, 9.0\}$, Baseline scale $w_{base} = 7.0$
\ENSURE Optimal guidance schedule $\mathbf{w}^* = (w^*_1, w^*_2, \dots, w^*_T)$
\STATE Initialize $\mathbf{w}^* = (w_{base}, w_{base}, \dots, w_{base})$ \COMMENT{Set baseline schedule}
\FOR{$t = 1$ \textbf{to} $T$}
    \STATE \COMMENT{Search for the optimal scale at current step $t$}
    \STATE $w^*_t = \arg\max_{w \in \mathcal{W}} \text{AverageMetric}\left( \text{Generate}(\mathcal{P}, \mathbf{w}_{tmp}) \right)$
    \STATE \textbf{where} $\mathbf{w}_{tmp} = (w^*_1, \dots, w^*_{t-1}, w, w_{base}, \dots, w_{base})$
    \STATE Update $\mathbf{w}^*$ with the newly found $w^*_t$
\ENDFOR
\STATE \textbf{Output:} Correlate each $w^*_t$ with the corresponding average ratio $r_t$ observed at step $t$ to analyze the trend in \Cref{fig:ratio_guidance_fit}.
\STATE \textbf{return} $\mathbf{w}^*$
\end{algorithmic}
\end{algorithm}

\section{Supplementary Results}

\subsection{Quantitative Evaluation via FID}
\label{app:exp_fid}

To rigorously assess the distributional similarity between generated images and real-world data, we evaluate \mtdbk using the Fr\'echet Inception Distance (FID) on 5,000 prompts from MS-COCO dataset. All samples are generated under the fixed seed. The results, summarized in \Cref{tab:fid_results}, demonstrate that \mtdbk significantly enhances the generative quality in low-step regimes. Specifically, at only $10$ sampling steps, \mtdbk achieves an FID of \textbf{32.05}, representing a substantial improvement over the standard CFG baseline ($63.62$). Notably, our $10$-step performance closely approaches the quality of $30$-step CFG ($24.80$), effectively bridging the gap between efficient sampling and high-fidelity generation. This confirms that \mtdbk still maintains an accurate generation trajectory even under aggressive acceleration.

\begin{table}[h]
    \centering
    \caption{\textbf{FID Results on MS-COCO.} We compare \mtdbk with standard CFG across different sampling steps. FID ($\downarrow$) measures the distributional distance to reference images (we use $50$-step CFG as reference). The results demonstrate that \mtdbk at $10$ steps significantly outperforms the $10$-step CFG baseline.}
    \label{tab:fid_results}
    \normalsize
    \begin{tabular}{lcc}
        \toprule
        Method & Sampling Steps & FID ($\downarrow$) \\
        \midrule
        CFG (Baseline) & 10 & 63.6170 \\
        CFG (Baseline) & 30 & 24.8043 \\
        \midrule
        \textbf{\mtdbk (Ours)} & \textbf{10} & \textbf{32.0545} \\
        \midrule
        CFG (Reference) & 50 & --- \\
        \bottomrule
    \end{tabular}
\end{table}

\subsection{Ablation Studies}
\label{app:exp_ablation}

\begin{figure}[h]
    \centering
    \begin{subfigure}{0.49\textwidth}
        \includegraphics[width=\linewidth]{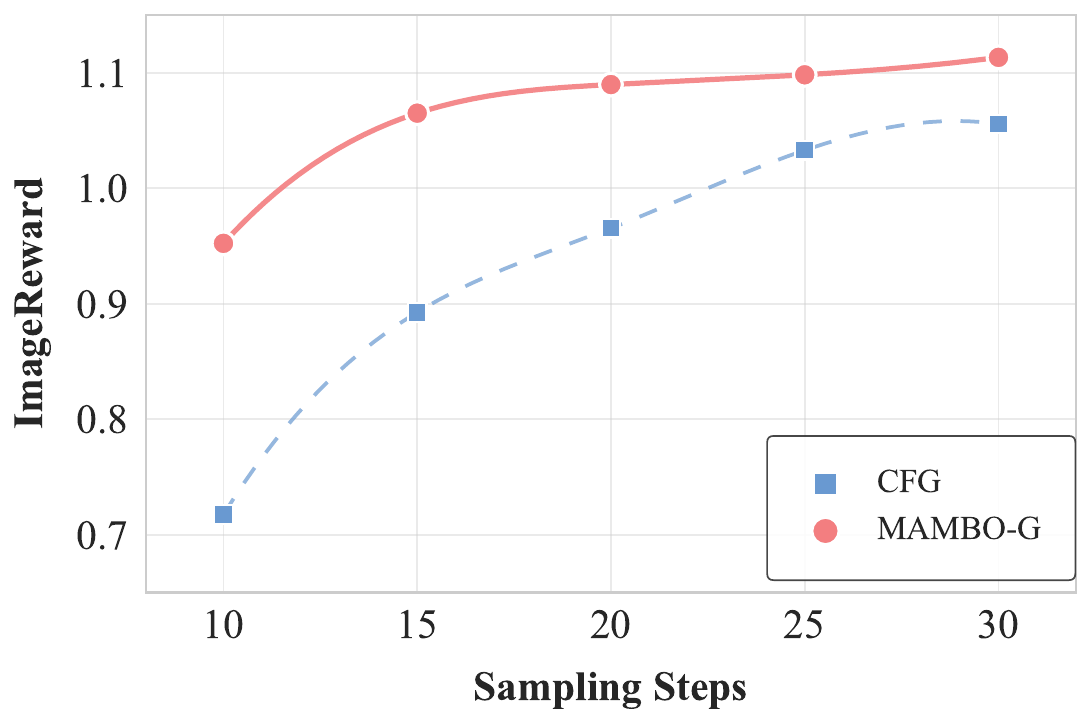}
        \captionsetup{justification=justified, singlelinecheck=true}
        \caption{SD3.5 UniPC, ImageReward.}
        \label{fig:sub1_1}
    \end{subfigure}
    \hfill
    \begin{subfigure}{0.49\textwidth}
        \includegraphics[width=\linewidth]{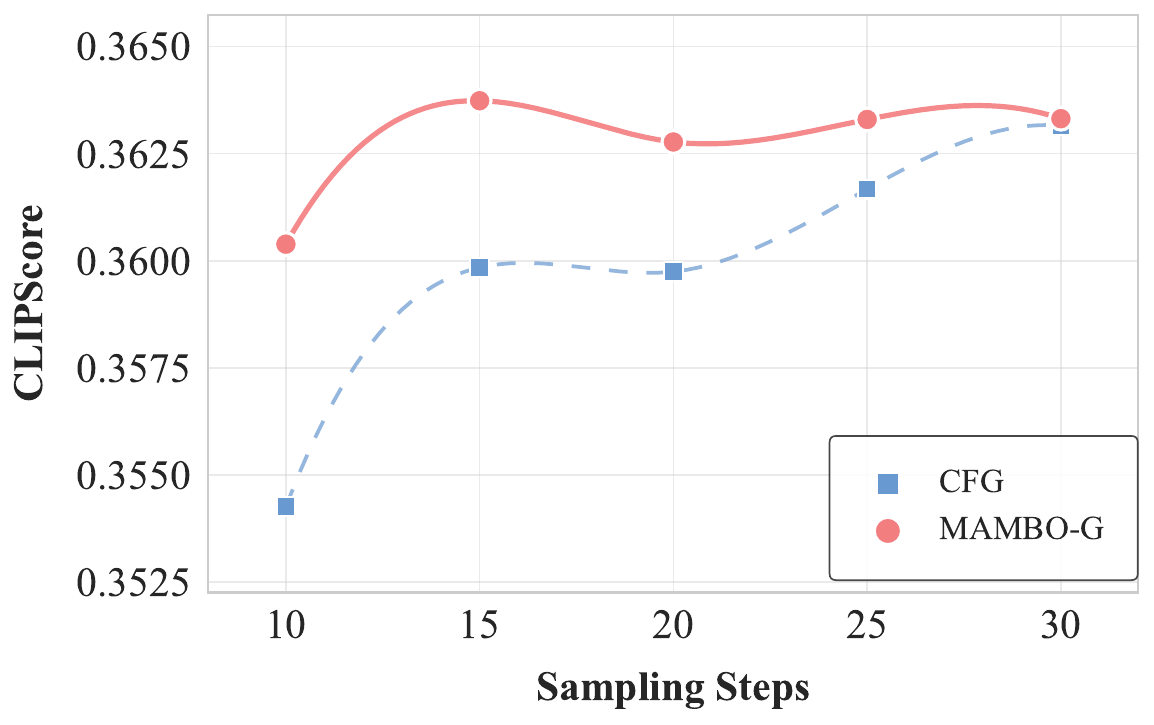}
        \captionsetup{justification=justified, singlelinecheck=true}
        \caption{SD3.5 UniPC, CLIPScore.}
        \label{fig:sub1_2}
    \end{subfigure}
    \captionsetup{justification=justified, singlelinecheck=false}
    \caption{Ablation studies of schedulers on UniPC. The comparative results present the consistent superiority of \mtdbk over CFG across different schedulers, further validating the wide-ranging adaptability of our method.}
    \label{fig:unipc_ablation}
\end{figure}

In this section, we present some supplementary results of our ablation studies in \Cref{subsec:ablation}, as shown in \Cref{tab:wmax_ablation,tab:alpha_ablation,fig:unipc_ablation}.
The detailed configurations are as follows:

\begin{itemize}

    \item \textbf{Ablation studies on $w_{\max}$ (\Cref{tab:wmax_ablation}): }
    We evaluate the sensitivity of $w_{\max}$ across SD3.5 and Qwen-Image models. Experiments are conducted on the first $100$ prompts from MS-COCO, following the randomization strategy defined in \Cref{subsec:txt2img_generation}. The hyperparameter $\alpha$ is fixed at $8$, while $w_{\max}$ is varied from $6$ to $16$. Each generation is performed with $10$ sampling steps.

    \item \textbf{Ablation studies on $\alpha$ (\Cref{tab:alpha_ablation}): }
    We evaluate the sensitivity of $\alpha$ across SD3.5 and Qwen-Image models. Experiments are conducted on the first $100$ prompts from MS-COCO, following the randomization strategy defined in \Cref{subsec:txt2img_generation}. The hyperparameter $w_{\max}$ is fixed at $10$, while $\alpha$ is varied from $6$ to $16$ to observe its impact on guidance damping. Each generation is performed with $10$ sampling steps.

    \item \textbf{Ablation studies on schedulers (\Cref{fig:unipc_ablation}): }
    We evaluate the robustness of \mtdbk on the UniPC scheduler. Experiments are conducted on the first $100$ prompts from MS-COCO, following the randomization strategy outlined in \Cref{subsec:txt2img_generation}. The hyperparameter settings remain consistent with those described in \Cref{subsec:txt2img_generation}. Each generation is performed with $10$ sampling steps.

\end{itemize}

\begin{table}[htbp]
\centering
\caption{Ablation studies on hyperparameter $w_{\max}$ (with fixed $\alpha=8$) across different models. The scores represent ImageReward, with the best performance in each row highlighted in \textbf{bold}.}
\label{tab:wmax_ablation}
\begin{tabular}{llcccccc}
\toprule
\multirow{2}{*}{\textbf{Model}} & \multirow{2}{*}{\textbf{Method}} & \multicolumn{6}{c}{\textbf{$w_{\max}$}} \\
\cmidrule(lr){3-8}
 & & \textbf{6} & \textbf{8} & \textbf{10} & \textbf{12} & \textbf{14} & \textbf{16} \\ 
\midrule
\multirow{2}{*}{SD3.5} & \mtdbk & 0.755 & 0.830 & 0.840 & \textbf{0.878} & 0.833 & 0.819 \\
 & \mtdbk + Rescale & 0.740 & 0.838 & 0.905 & 0.895 & 0.903 & \textbf{0.912} \\ 
\midrule
\multirow{2}{*}{Qwen-Image} & \mtdbk & \textbf{1.112} & 1.094 & 1.081 & 0.946 & 0.899 & 0.780 \\
 & \mtdbk + Rescale & \textbf{1.126} & 1.095 & 1.122 & 1.093 & 1.058 & 1.042 \\ 
\bottomrule
\end{tabular}
\end{table}

\begin{table}[htbp]
\centering
\caption{Ablation studies on hyperparameter $\alpha$ (with fixed $w_{\max}=10$) across different models. The scores represent ImageReward, with the best performance in each row highlighted in \textbf{bold}.}
\label{tab:alpha_ablation}
\begin{tabular}{llcccccc}
\toprule
\multirow{2}{*}{\textbf{Model}} & \multirow{2}{*}{\textbf{Method}} & \multicolumn{6}{c}{\textbf{$\alpha$}} \\
\cmidrule(lr){3-8}
 & & \textbf{6} & \textbf{8} & \textbf{10} & \textbf{12} & \textbf{14} & \textbf{16} \\ 
\midrule
\multirow{2}{*}{SD3.5} & \mtdbk & 0.827 & 0.840 & \textbf{0.857} & 0.817 & 0.785 & 0.774 \\
 & \mtdbk + Rescale & 0.875 & \textbf{0.905} & 0.867 & 0.818 & 0.808 & 0.768 \\ 
\midrule
\multirow{2}{*}{Qwen-Image} & \mtdbk & 0.830 & 1.081 & 1.156 & \textbf{1.161} & 1.132 & 1.122 \\
 & \mtdbk + Rescale & 1.010 & 1.122 & \textbf{1.155} & 1.142 & 1.136 & 1.130 \\ 
\bottomrule
\end{tabular}
\end{table}

\end{document}